# Recent Advances and Clinical Applications of Deep Learning in Medical Image Analysis

Xuxin Chen[a], Ximin Wang[b], Ke Zhang[a], Kar-Ming Fung[c], Theresa C. Thai[d], Kathleen Moore[e], Robert S. Mannel[e], Hong Liu[a], Bin Zheng[a], Yuchen Qiu[a*]

[a] School of Electrical and Computer Engineering, University of Oklahoma, Norman, OK, USA 73019
[b] School of Information Science and Technology, ShanghaiTech University, Shanghai, China 201210
[c] Department of Pathology, University of Oklahoma Health Sciences Center, Oklahoma City, OK, USA 73104
[d] Department of Radiology, University of Oklahoma Health Sciences Center, Oklahoma City, OK, USA 73104
[e] Department of Obstetrics and Gynecology, University of Oklahoma Health Sciences Center, Oklahoma City, OK, USA 73104

## ABSTRACT

Deep learning has received extensive research interest in developing new medical image processing algorithms, and deep learning based models have been remarkably successful in a variety of medical imaging tasks to support disease detection and diagnosis. Despite the success, the further improvement of deep learning models in medical image analysis is majorly bottlenecked by the lack of large-sized and well-annotated datasets. In the past five years, many studies have focused on addressing this challenge. In this paper, we reviewed and summarized these recent studies to provide a comprehensive overview of applying deep learning methods in various medical image analysis tasks. Especially, we emphasize the latest progress and contributions of state-of-the-art unsupervised and semi-supervised deep learning in medical image analysis, which are summarized based on different application scenarios, including classification, segmentation, detection, and image registration. We also discuss the major technical challenges and suggest the possible solutions in future research efforts.

## 1. INTRODUCTION

In current clinical practice, accuracy of detection and diagnosis of cancers and/or many other diseases depends on the expertise of individual clinicians (e.g., radiologists, pathologists) (Kruger et al., 1972), which results in large inter-reader variability in reading and interpreting medical images. In order to address and overcome this clinical challenge, many computer-aided detection and diagnosis (CAD) schemes have been developed and tested, aiming to help clinicians more efficiently read medical images and make the diagnostic decision in a more accurate and objective manner. The scientific rationale of this approach is that using computer-aided quantitative image feature analysis can help overcome many negative factors in the clinical practice, including the wide variations in expertise of the clinicians, potential fatigue of human experts, and lack of sufficient medical resources.

Although early CAD schemes have been developed in 1970s (Meyers et al., 1964; Kruger et al., 1972; Sezaki and Ukena, 1973), progress of the CAD schemes accelerates since the middle of 1990s (Doi et al., 1999), due to the development and integration of more advanced machine learning methods or models into CAD schemes. For conventional CAD schemes, a common developing procedure consists of three steps: target segmentation, feature computation, and disease classification. For example, Shi et al. (2008) developed a CAD scheme to achieve mass classification on digital mammograms. The ROIs containing the target masses were first segmented from the background using a modified active contour algorithm (Sahiner et al., 2001). Then a large number of image features were applied to quantify the lesion characteristics in size, morphology, margin geometry, texture, and etc. Thus the raw pixel data was converted into a vector of representative features. Finally, a LDA (linear discrimination analysis) based classification model was applied on the feature vector to identify the mass malignancy.

[*] Corresponding author: qiuyuchen@ou.edu

As a comparison, for deep learning based models, hidden patterns inside ROIs are progressively identified and learned by the hierarchical architecture of deep neural networks (LeCun et al., 2015). During this process, important properties of the input image will be gradually identified and amplified for certain tasks (e.g. classification, detection), while irrelevant features will be attenuated and filtered out. For instance, an MRI image depicting suspicious liver lesions comes with a pixel array (Hamm et al., 2019), and each entry is used as one input feature of the deep learning model. The first several layers of the model may initially obtain some basic lesion information, such as tumor shape, location, and orientation. The next batch of layers may identity and keep the features consistently related to lesion malignancy (e.g. shape, edge irregularity), while ignoring irrelevant variations (e.g. location). The relevant features will be further processed and assembled by subsequent higher layers in a more abstract manner. When increasing the number of layers, a higher level of feature representations can be achieved. Through the entire procedure, important features hidden inside the raw image are recognized by a general neural network based model in a self-taught manner, and thus the manual feature development is not needed.

Due to its huge advantage, deep learning related methods have become the mainstream technology in the CAD field and have been widely applied in a variety of tasks, such as disease classification (Li et al., 2020a; Shorfuzzaman and Hossain, 2021; Zhang et al., 2020a; Frid-Adar et al., 2018a; Kumar et al., 2016; Kumar et al., 2017), ROI segmentation (Alom et al., 2018; Yu et al., 2019; Fan et al., 2020), medical object detection (Rijthoven et al., 2018; Mei et al., 2021; Nair et al., 2020; Zheng et al., 2015), and image registration (Simonovsky et al., 2016; Sokooti et al., 2017; Balakrishnan et al., 2018). Among different kinds of deep learning techniques, supervised learning was first adopted in medical image analysis. Although it has been successfully utilized in many applications (Esteva et al., 2017; Long et al., 2017), further deployment of supervised models in many scenarios is majorly hurdled by the limited size of most medical datasets. As compared to regular datasets in computer vision, a medical image dataset usually contains relatively small amounts of images (less than 10,000), and in many cases, only a small percentage of images are annotated by experts. To overcome this limitation, unsupervised and semi-supervised learning methods have received extensive attention in the past three years, which are able to (1) generate more labeled images for model optimization, (2) learn meaningful hidden patterns from unlabeled image data, and (3) generate pseudo labels for the unlabeled data.

There already exist a number of excellent review articles that summarized deep learning applications in medical image analysis. Litjens et al. (2017) and Shen et al. (2017) reviewed relatively early deep learning techniques, which are mainly based on supervised methods. More recently, Yi et al. (2019) and Kazeminia et al. (2020) reviewed the applications of generative adversarial networks across different medical imaging tasks. Cheplygina et al. (2019) surveyed on how to use semi-supervised learning and multiple instance learning in diagnosis or segmentation tasks. Tajbakhsh et al. (2020) investigated a variety of methods to deal with dataset limitations (e.g., scarce or weak annotations) specifically in medical image segmentation. In contrast, a major goal of this paper is to shed light on how the medical image analysis field, which is often bottlenecked by limited, annotated data, can potentially benefit from some latest trends of deep learning. Our survey distinguishes itself from recent review papers with two characteristics – being comprehensive and technically oriented. "Comprehensive" is reflected in three aspects. First, we highlight the applications of a broad range of promising approaches falling in the "not-so-supervised" category, including self-supervised, unsupervised, semi-supervised learning; meanwhile, we do not ignore the continuing importance of supervised methods. Second, rather than covering only a specific task, we introduce the applications of the above learning approaches in four classical medical image analysis tasks (classification, segmentation, detection, and registration). Especially, we discussed the deep learning based object detection in detail, which is rarely mentioned in recent review papers (after 2019). We focused on the applications of chest X-ray, mammogram, CT, and MRI images. All these types of the images have many common characteristics, which are interpreted by physicians at the same department (Radiology). We also mentioned some general methods which were applied on other image domains (e.g. histopathology) but have potential to be used in radiological or MRI images. Third, state-of-the-art architectures/models for these tasks are explained. For example, we summarized how to adapt *Transformers* from natural language processing for medical image segmentation, which has not been mentioned by existing review papers to the best of our knowledge. In terms of "technically oriented", we review the recent advances of not-so-supervised approaches in detail. In particular, self-supervised learning is quickly emerging as a promising direction but yet systematically reviewed in the context of medical vision. A wide audience may benefit from this survey, including researchers with deep learning, artificial intelligence and big data expertise, and clinicians/medical researchers.

This survey is presented as follows (Figure 1): Section 2 provides an in-depth overview of recent advances in deep learning, with a focus on unsupervised and semi-supervised approaches. In addition, three important strategies for performance enhancement, including attention mechanisms, domain knowledge, and uncertainty estimation, are also discussed. Section 3 summarizes the major contributions of applying deep learning techniques in four main tasks: classification, segmentation, detection, and registration. Section 4 discusses challenges for further improving the model and suggests possible perspectives on future research directions toward large scale applications of deep learning based medical image analysis models.

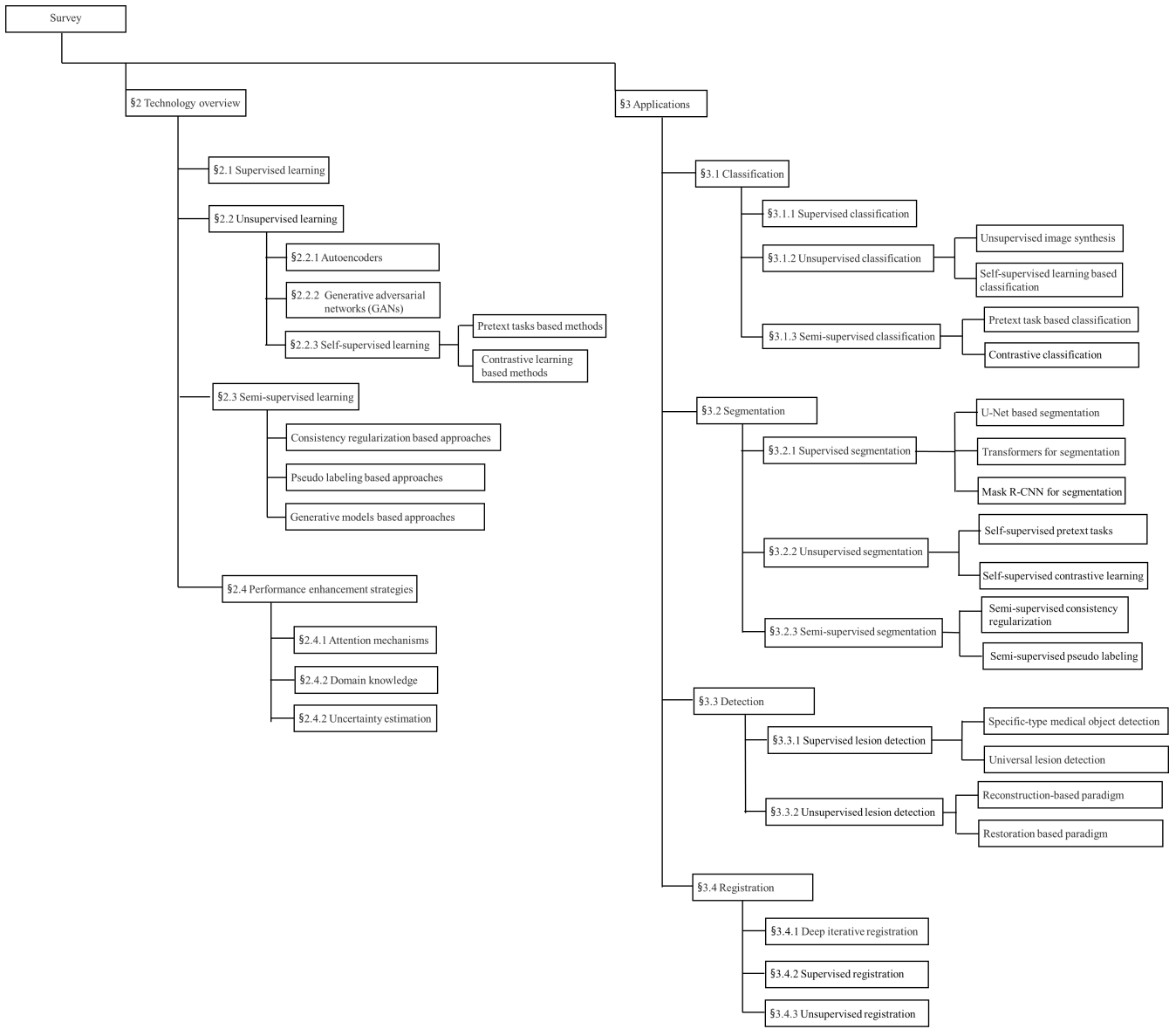

Figure 1. The overall structure of this survey.

## 2. OVERVIEW OF DEEP LEARNING METHODS

Depending on whether labels of the training dataset are present, deep learning can be roughly divided into supervised, unsupervised, and semi-supervised learning. In **supervised** learning, all training images are labeled, and the model is optimized using the image-label pairs. For each testing image, the optimized model will generate a likelihood score to predict its class label (LeCun et al., 2015). For **unsupervised** learning, the model will analyze and learn the underlying patterns or hidden data structures without labels. If only a small portion of training data is labeled, the model learns input-output relationship from the labeled data, and the model will be strengthened by learning semantic and fine-grained features from the unlabeled data. This type of learning approach is defined as **semi-supervised** learning (van Engelen and Hoos, 2020). In this section, we briefly mentioned supervised learning at the beginning, and then majorly reviewed the recent advances of unsupervised learning and semi-supervised learning, which can facilitate performing medical image tasks with limited annotated data. Popular frameworks for these two types of learning paradigms will be introduced accordingly. In the end, we summarize three general strategies that can be combined with different learning paradigms for better performance in medical image analysis, including attention mechanisms, domain knowledge, and uncertainty estimation.

### 2.1. Supervised learning

Convolutional neural networks (CNNs) are a widely used deep learning architecture in medical image analysis (Anwar et al., 2018). CNNs are mainly composed of convolutional layers and pooling layers. Figure 2 shows a simple CNN in the context of medical image classification task. The CNN directly takes an image as input, and transforms it via convolutional layers, pooling layers, and fully connected layers, and finally outputs a class-based likelihood of that image.

At each convolutional layer $l$, a bunch of kernels $W = \{W_1, ..., W_k\}$ are used to extract features from the input image, and biases $b = \{b_1, ..., b_k\}$ are added, generating new feature maps $W_i^l x_i^l + b_i^l$. Then a non-linear transform, an activation function $\sigma(.)$, is applied resulting in $x_k^{l+1} = \sigma(W_i^l x_i^l + b_i^l)$ as the input of the next layer. After the convolutional layer, a pooling layer is incorporated to reduce the dimension of feature maps, thus reducing the number of parameters. Average pooling and maximum pooling are two common pooling operations. The above process is repeated for the rest layers. At the end of the network, fully connected layers are usually employed to produce the probability distribution over classes via a sigmoid or softmax function. The predicted probability distribution gives a label $\hat{y}$ for each input instance so that a loss function $L(\hat{y}, y)$ can be calculated, where $y$ is the real label. Parameters of the network are iteratively optimized by minimizing the loss function.

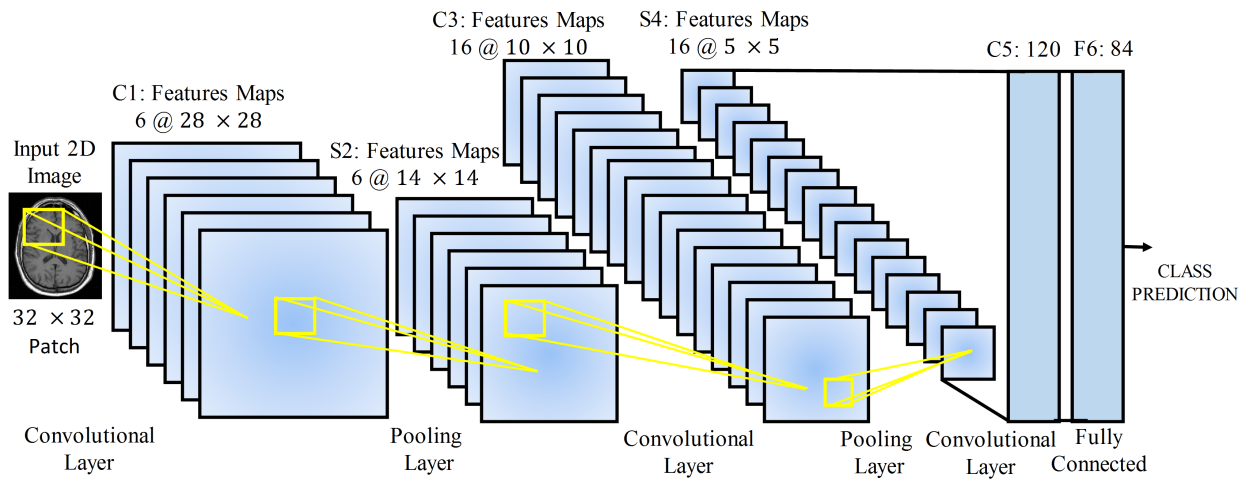

Figure 2. A simple CNN for disease classification from MRI images (Anwar et al., 2018).

## 2.2. Unsupervised learning

### 2.2.1. Autoencoders

Autoencoders are widely applied in dimensionality reduction and feature learning (Hinton and Salakhutdinov, 2006). The simplest autoencoder, initially known as auto-associator (Bourlard and Kamp, 1988), is a neural network with only one hidden layer that learns a latent feature representation of the input data by minimizing a reconstruction loss between the input and its reconstruction from the latent representation. The shallow structure of simple autoencoders limits their representation power, but deeper autoencoders with more hidden layers can improve the representation. By stacking multiple auto-encoders and optimizing them in a greedy layer-wise manner , deep autoencoders or Stacked Autoencoders (SAEs) can learn more complicated non-linear patterns than shallow ones and thus generalize better outside training data (Bengio et al., 2007). SAEs consist of an encoder network and a decoder network, which are typically symmetrical to each other. To further force models to learn useful latent representations with desirable characteristics, regularization terms such as sparsity constraints in Sparse Autoencoders (Ranzato et al., 2007) can be added to the original reconstruction loss. Other regularized autoencoders include the Denoising Autoencoder (Vincent et al., 2010) and Contractive Autoencoder (Rifai et al., 2011), both designed to be insensitive to input perturbations.

Unlike the above classic autoencoders, variational autoencoder (VAE) (Kingma and Welling, 2013) works in a probablistic manner to learn mappings between the observed data space $x \in R^m$ and latent space $z \in R^n$ ($m \gg n$). As a latent variable model, VAE formulates this problem as maximizing the log-likelihood of the observed samples $\log p(x) = \log \int p(x|z)p(z)\, dz$, where $p(x|z)$ can be easily modeled using a neural network, and $p(z)$ is a prior distribution (such as Gaussian) over the latent space. However, the integral is intractable because it is impossible to sample the full latent space. As a result, the posterior distribution $p(z|x)$ also becomes intractable according to Bayes rule. To solve the intractability issue, the authors of VAE propose that in addition to modeling $p(x|z)$ using the decoder, the encoder learns $q(z|x)$ that approximates the unknown posterior distribution. Ultimately a tractable lower bound also termed as evidence lower bound (EBLO), can be derived for $\log p(x)$.

$$\log p(x) \geq ELBO = E_{q(z|x)}[\log p(x|z)] - KL[q(z|x)||p(z)],$$

where $KL$ stands for the Kullback-Leibler divergence. The first term can be understood as a reconstruction loss measuring the similarity between the input image and its counterpart reconstructed from the latent representation. The second term computes the divergence between the approximated posterior and Gaussian prior.

Later different VAE extensions were proposed to learn more complicated representations. Although the probabilistic working mechanism allows its decoder to generate new data, VAE cannot specifically control the data generation process. Sohn et al. (2015) proposed the so-called conditional VAE (CVAE), where probabilistic distributions learnt by the encoder and decoder are both conditioned using external information (e.g. image classes). This enables VAE to generate structured output representations. Another line of research explores imposing more complex priors on the latent space. For example, Dilokthanakul et al. (2016) presented Gaussian Mixture VAE (GMVAE) that uses a mixture of Gaussians as prior to obtain higher modeling compacity in latent space. We refer readers to a recent paper (Kingma and Welling, 2019) for more details of VAE and its extensions.

### 2.2.2. Generative adversarial networks (GANs)

Generative adversarial networks (GANs) are a class of deep nets for generative modeling first proposed by Goodfellow et al. (2014). For this architecture, a framework for estimating generative models is designed to directly draw samples from the desired underlying data distribution without the need to explicitly define a probability distribution. It consists of two models: a generator G and a discriminator D. The generative model G takes as input a random noise vector $z$ sampled from a prior distribution $P_z(z)$ , often either a Gaussian or a uniform distribution, and then maps $z$ to data space as $\boldsymbol{G}(z, \theta_g)$, where $\boldsymbol{G}$ is a neural network with parameters $\theta_g$. The fake samples denoted as $\boldsymbol{G}(z)$ or $x_g$ are expected to resemble real samples from the training data $P_r(x)$, and these two types of samples are sent into D. The discriminator, a second neural network parameterized by $\theta_d$, outputs the probability $D(x, \theta_d)$ that a sample comes from the training data rather than G. The training procedure is like playing a minimax two-player game. The discriminative network D is optimized to maximize the log likelihood of assigning correct labels to fake samples and real samples, while the generative model G is trained to maximize the log likelihood of D making a mistake. Through the adversarial process, G is desired to gradually estimate the underlying data distribution and generate realistic samples.

Based on the vanilla GAN, the performance was improved in the following two directions: 1) different loss (objective) functions, and 2) conditional settings. For the first direction, Wasserstein GAN (WGAN) is a typical example. In WGAN, Earth-Mover (EM) distance or Wasserstein-1, commonly known as the Wasserstein distance, was proposed to replace the Jensen–Shannon (JS) divergence in original Vanilla GAN and measure the distance between the real and synthetic data distribution (Arjovsky et al., 2017). The critic of WGAN has the advantage to provide useful gradients information where JS divergence saturates and results in vanishing gradients. WGAN could also improve the stability of learning and alleviate problems like mode collapse.

An unconditional generative model cannot explicitly control the modes of data being synthesized. To guide the data generation process, the conditional GAN (cGAN) is constructed by conditioning its generator and discriminator with additional information (i.e., the class labels) (Mirza and Osindero, 2014). Specifically, the noise vector $z$ and class label $c$ are jointly provided to G; the real/ fake data and class label $c$ are together presented as the inputs of D. The conditional information can also be images or other attributes, not limited to class labels. Further, the auxiliary classifier GAN (ACGAN) presents another strategy to employ label conditioning to improve image synthesis (Odena et al., 2017). Unlike the discriminator of cGAN, D in ACGAN is no longer provided with the class conditional information. Apart from separating real and fake images, D is also tasked with reconstructing class labels. When forcing D to perform the additional classification task, ACGAN can generate high-quality images easily.

### 2.2.3. Self-supervised learning

In the past few years, unsupervised representation learning has gained huge success in natural language processing (NLP), where massive unlabeled data is available for pre-training models (e.g. BERT, Kenton and Toutanova, 2019) and learning useful feature representations. Then the feature representations are fine-tuned in downstream tasks such as question answering, natural language inference, and text summarization. In computer vision, researchers have explored a similar pipeline – models are first trained to learn rich and meaningful feature representations from the raw unlabeled image data in an unsupervised manner, and then the feature representations are fine-tuned in a wide variety of downstream tasks with labeled data, such as classification, object detection, instance segmentation, etc. However, this practice was not as successful as in NLP for quite a long time, and instead supervised pre-training has been the dominant strategy. Interestingly, we find this situation is changing toward the opposite direction in recent two years, as more and more studies report a higher performance of self-supervised pre-training than supervised pre-training.

In recent literature, the term *self-supervised learning* is used interchangeably with *unsupervised learning*; more accurately, self-supervised learning actually refers to a form of deep unsupervised learning, where inputs and labels are created from unlabeled data itself without external supervision. One important motivation behind this technology is to avoid supervised tasks that are often expensive and time-consuming, due to the need to establish new labeled datasets or acquire high-quality annotations in certain fields like medicine. Despite the scarcity and high cost of labeled data, there usually exist large amounts of cheap unlabeled data remaining unexploited in many fields. The unlabeled data is likely to contain valuable information that is either weak or not present in labeled data. Self-supervised learning can leverage the power of unlabeled data to improve both the performance and efficiency in supervised tasks. Since self-supervised learning touches upon vaster data than supervised learning, features learnt in a self-supervised manner can potentially better generalize in the real world. Self supervision can be created in two ways: **pretext tasks** based methods and **contrastive learning** based methods. Since the contrastive learning based methods have received broader attention in very recent years, we will highlight more works in this direction.

**Pretext task** is designed to learn representative features for downstream tasks, but the pretext itself is not of the true interest (He et al., 2020). The pretext tasks learn representations by hiding certain information (e.g., channel, patches, etc.) for each input image, and then predict the missing information from the image's remaining parts. Examples include image inpainting (Pathak et al., 2016), colorization (Zhang et al., 2016), relative patch prediction (Doersch et al., 2015), jigsaw puzzles (Noroozi and Favaro, 2016), rotation (Gidaris et al., 2018), etc. However, the learnt representations' generalizability is heavily dependent on the quality of hand-crafted pretext tasks (Chen et al., 2020a).

**Contrastive learning** relies on the so-called contrastive loss, which can date back to at least (Hadsell et al., 2006; Chopra et al., 2005a). Later a number of variants of this contrastive loss were used (Oord et al., 2018; Chen et al., 2020a; Chaitanya et al., 2020). In essence, the original loss and its later versions all enforce a similarity metric to be maximized for positive (similar) pairs and be minimized for negative (dissimilar) pairs, so that the model can learn discriminative features. In the following we will introduce two representative frameworks for contrastive learning, namely Momentum Contrast (MoCo) (He et al., 2020) and SimCLR (Chen et al., 2020a).

MoCo formulates contrastive learning as a dictionary look-up problem, which requires an encoded query to be similar to its matching key. As shown in Figure 3 (a), given an image $x$, an encoder encodes the image resulting in a feature vector, which is used as a *query* ($q$). Likewise, with another encoder the dictionary can be built up by the features $\{k_0, k_1, k_2, ...\}$, also known as *keys*, from a large set of image samples $\{x_0, x_1, x_2, ...\}$. In MoCo, the encoded query $q$ and a key are considered similar if they come from different crops of the same image. Suppose there exists a single dictionary key ($k^+$) that matches with $q$, then these two items are regarded as a positive pair, whereas the rest keys in the dictionary are considered negative. The authors compute the loss function of a positive pair using InfoNCE (Oord et al., 2018) as follows:

$$L_q = -\log \frac{\exp(q.k^+/\tau)}{\sum_{i=0}^{K} \exp(q.k_i/\tau)}$$

Established from a sampled subset of all images, a large dictionary is important for good accuracy. To make the dictionary large, the authors maintain the feature representations from previous image batches as a *queue:* new keys are enqueued with old keys dequeued. Therefore, the dictionary consists of encoded representations from the current and previous batches. This, however, could lead to a rapidly updated key encoder, rendering the dictionary keys inconsistent, i.e., their comparisons to the encoded query are not consistent. The authors thus propose using momentum update on the key encoder to avoid rapid changes. This key encoder is referred as the momentum encoder.

SimCLR is another popular framework for contrastive learning. In this framework, two augmented images are considered a postitive pair if they derive from the same example; if not, they are a negative pair. The agreement of feature representations from of postive image pairs are maximized. As shown in Figure 3 (b), SimCLR consists of four components: (1) stochastic image augmentation; (2) encoder networks ($f(.)$) extracting feature representations from augmented images; (3) a small neural network (multilayer perceptron (MLP) projection head) ($g(.)$) that maps the feature representations to a lower-dimensional space; and (4) contrastive loss computation. The third component differs SimCLR from its predecessors. Previous frameworks like MoCo compute the feature representations directly rather than first mapping them to a lower-dimensional space. This component is further proven important in achieving satisfactory results, as demonstrated in MoCo v2 (Chen et al., 2020b).

Note that since self-supervised contrastive learning is very new, wide applications of recent advances such as MoCo and SimCLR in the medical image analysis field have yet been established at the time of this writing. Nonetheless, considering the promising results of self-supervised learning reported in the existing literature, we anticipate studies applying this new technology to analyze medical images are likely to explode soon. Also, self-supervised pre-training has great potential to be a strong alternative of supervised pre-training.

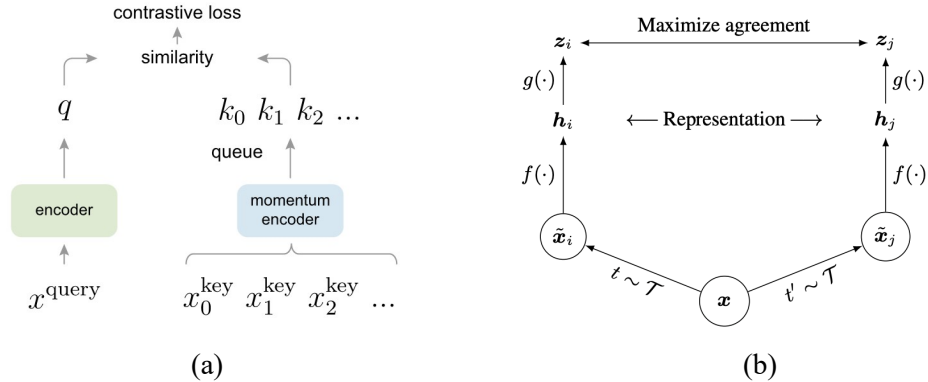

Figure 3. (a) MoCo (He et al., 2020); (b) SimCLR (Chen et al., 2020a).

## 2.3. Semi-supervised learning

Different from unsupervised learning that can work just on unlabeled data to learn meaningful representations, semi-supervised learning (SSL) combines labeled and unlabeled data during model training. Especially, SSL applies to the scenario where limited labeled data and large-scale but unlabeled data are available. These two types of data should be relevant, so that the additional information carried by unlabeled data could be useful in compensating the labeled data. It is reasonable to expect that unlabeled data would lead to an average performance boost – probably the more the better for performing tasks with only limited labeled data. In fact, this goal has been explored for several decades, and the 1990s already witnessed a rising interest of applying SSL methods in text classification. The Semi-Supervised Learning book (Chapelle et al., 2009) is a good source for readers to grasp the connection of SSL to classic machine learning algorithms. Interestingly, despite its potential positive value, the authors present empirical findings that unlabeled data sometimes deteriorates the performance. However, this empirical finding seems to be experiencing changes in recent literature of deep learning – an increasing number of works, mostly from the computer vision field, have reported that deep semi-supervised approaches generally perform better than high-quality supervised baselines (Ouali et al., 2020). Even when varying the amount of labeled and unlabeled data, a consistent performance improvement can still be observed. At the same time, deep semi-supervised learning has been successfully applied in the medical image analysis field to reduce annotation cost and achieve better performance. We divide popular SSL methods into three groups: (1) consistency regularization based approaches; (2) pseudo labeling based approaches; (3) generative models based approaches.

Methods in the first category share one same idea that the prediction for an unlabeled example should not change significantly if some perturbations (e.g., adding noise, data augmentation) are applied. The loss function of an SSL model generally consists of two parts. More concretely, given an unlabeled data example $x$ and its perturbed version $\hat{x}$, the SSL model outputs logits $f_\theta(x)$ and $f_\theta(\hat{x})$. On the unlabeled data, the objective is to give consistent predictions by minimizing the mean squared error $d(f_\theta(x), f_\theta(\hat{x}))$, and this leads to the consistency (unsupervised) loss $L_u$ on unlabeled data. On the labeled data, a cross entropy supervised loss $L_s$ is computed. Example SSL models that are regularized by consistency constraints include Ladder Networks (Rasmus et al., 2015), $\Pi$-Model (Laine and Aila, 2017), and Temporal Ensembling (Laine and Aila, 2017). A more recent example is the *Mean Teacher* paradigm (Tarvainen and Valpola, 2017), composed of a teacher model and a student model (Figure 4). The student model is optimized by minimizing $L_u$ on unlabeled data and $L_s$ on labeled data; as an Exponential Moving Average (EMA) of the student model, the teacher model is used to guide the student model for consistency training. Most recently, several works such as unsupervised data augmentation (UDA) (Xie et al., 2020) and MixMatch (Berthelot et al., 2019) have brought the performance of SSL to a new level.

For pseudo labeling (Lee, 2013), an SSL model itself generates pseudo annotations for unlabeled examples; the pseudo-labeled examples are used jointly with labeled examples to train the SSL model. This process is iterated for several times, during which the quality of pseudo labels and the model's performance get enhanced. The naïve pseudo-labeling process can be combined with Mixup augmentation (Zhang et al., 2018a) to further improve SSL model's performance (Arazo et al., 2020). Pseudo labeling also works well with multi-view co-training (Qiao et al., 2018). For each view of the labeled examples, co-training learns a separate classifier, and then the classifier is used to generate pseudo labels for the unlabeled data; co-training maximizes the agreement of assigning pseudo annotations among each view of unlabeled examples.

For methods in the third category, semi-supervised generative models such as GANs and VAEs put more focus on solving target tasks (e.g., classification) than just generating high-fidelity samples. Here we illustrate the mechanism of semi-supervised GAN for brevity. One simple way to adapt GAN to semi-supervised settings is by modifying the discriminator to perform additional tasks. For example, in the task of image classification, Salimans et al. (2016) and Odena (2016) changed the discriminator in DCGAN by forcing it to serve as a classifier. For an unlabeled image, the discriminator functions as in the vanilla GAN, providing a probability of the input image being real; for a labeled image, the discriminator predicts its class besides generating a realness probability. However, Li et al. (2017) demonstrated that the optimal performance of the two tasks may not be achieved at the same time by a single discriminator. Thus, they introduced an additional classifier that is independent from the generator and discriminator. This new architecture composed of three components is called Triple-GAN.

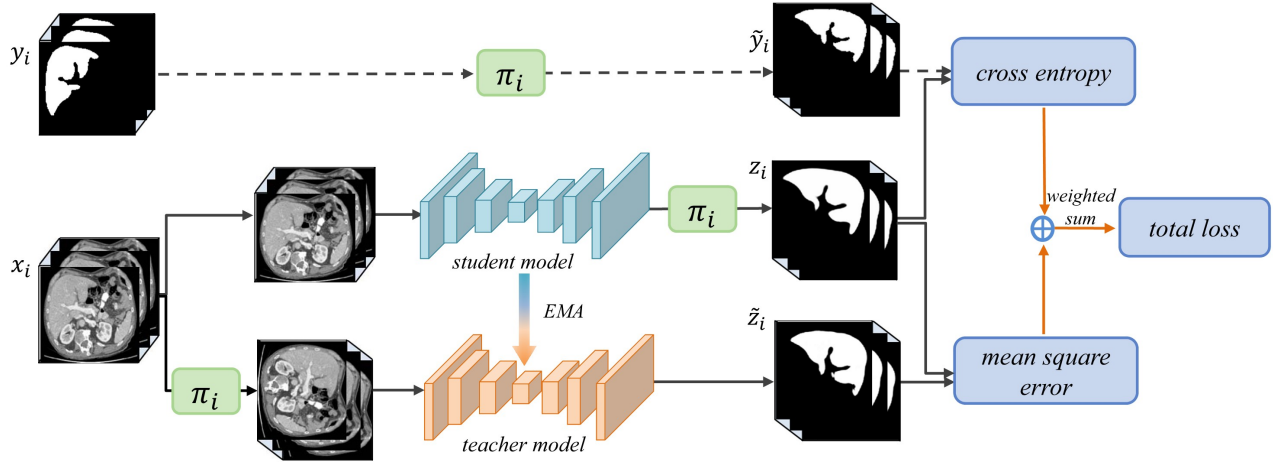

Figure 4. Mean Teacher model application in medical image analysis (Li et al., 2020b). $\pi_i$ refers to the transformation operations, including rotation, flipping, and scaling. $z_i$ and $\tilde{z}_i$ are network outputs.

## 2.4. Strategies for performance enhancement

### 2.4.1. Attention mechanisms

Attention originates from primates' visual processing mechanism that selects a subset of pertinent sensory information, rather than using all available information for complex scene analysis (Itti et al., 1998). Inspired by this idea of focusing on specific parts of inputs, deep learning researchers have integrated attention into developing advanced models in different fields. Attention-based models have achieved huge success in fields related to natural language processing (NLP), such as machine translation (Bahdanau et al., 2015; Vaswani et al., 2017) and image captioning (Xu et al., 2015; You et al., 2016; Anderson et al., 2018). One prominent example is the *Transformer* architecture that solely relies on self-attention to capture global dependencies between input and output, without requiring sequential computation (Vaswani et al., 2017). Attention mechanisms have also become popular in computer vision tasks, such as natural image classification (Wang et al., 2017; Woo et al., 2018; Jetley et al., 2018), segmentation (Chen et al., 2016; Ren and Zemel, 2017), etc. When processing images, attention modules can adaptively learn "what" and "where" to attend so that model predictions are conditioned on the most relevant image regions and features. Based on how attended locations in an image are selected, attention mechanisms can be roughly divided into two categories, namely soft and hard attention. The former deterministically learns a weighted average of features at all locations, whereas the latter stochastically samples one subset of feature locations to attend (Cho et al., 2015). Since hard attention is not differentiable, soft attention, despite being more computationally expensive, has received more research efforts. Following this differentiable mechanism, different types of attention have been further developed, such as (1) spatial attention (Jaderberg et al., 2015), (2) channel attention (Hu et al., 2018a), (3) combination of spatial and channel-wise attention (Wang et al., 2017; Woo et al., 2018), and (4) self attention (Wang et al., 2018). Readers are referred to the excellent review by Chaudhari et al. (2021) for more details of attention mechanisms.

### 2.4.2. Domain knowledge

Most well-established deep learning models, originally designed to analyze natural images, are likely to produce only suboptimal results when directly applied to medical image tasks (Zhang et al., 2020a). This is because natural and medical images are very different in nature. First, medical images usually exhibit high inter-class similarity, so one major challenge lies in the extraction of fine-grained visual features to understand subtle differences that are important to making correct predictions. Second, typical medical image datasets are much smaller than benchmark natural datasets that contain images ranging from tens of thousands to millions. This hinders models with high complexity in computer vision from being directly applied in the medical domain. Therefore, how to customize models for medical image analysis remains an important issue. One possible solution is to integrate proper domain knowledge or task-specific properties, which has proven beneficial to facilitate learning useful feature representations and reducing model complexity in the medical imaging context. In this review paper, we will mention a variety of domain knowledge, such as anatomical information in MRI and CT images (Zhou et al., 2021; Zhou et al., 2019a), 3D spatial context information from volumetric images (Zhang et al., 2017; Zhuang et al., 2019; Zhu et al., 2020a), multi-instance data from the same patient (Azizi et al., 2021), patient metadata (Vu et al., 2021), radiomic features (Shorfuzzaman and Hossain, 2021), text reports accompanying images (Zhang et al., 2020a), etc. Readers interested in a comprehensive review of how to integrate medical domain knowledge into network designing can refer to the work of Xie et al. (2021a).

### 2.4.3. Uncertainty estimation

Reliability is of critical concern when it comes to clinical settings with high-safety requirements (e.g. cancer diagnosis). Model predictions are easily subject to factors such as data noise and inference errors, so it is desirable to quantify uncertainty and make the results trustworthy (Abdar et al., 2021). Commonly used techniques for uncertainty estimation include Bayesian approximation (Gal and Ghahramani, 2016) and model ensemble (Lakshminarayanan et al., 2017). Bayesian approaches like Monte Carlo dropout (MC-dropout) (Gal and Ghahramani, 2016) revolve around approximating the posterior distribution over neural networks' parameters. Ensemble techniques combine multiple models to measure uncertainty. Readers interested in uncertainty estimation are referred to the comprehensive review by Abdar et al. (2021).

# 3. DEEP LEARNING APPLICATIONS

## 3.1. Classification

Medical image classification is the goal of computer-aided diagnosis (CADx), which aims at either distinguishing malignant lesions from benign ones or identifying certain diseases from input images (Shen et al., 2017; van Ginneken et al., 2011). Deep learning based CADx schemes have received huge success over the last decade. However, deep neural nets generally depend on sufficient annotated images to ensure good performance, and this requirement may not be easily satisfied by many medical image datasets. To alleviate the lack of large annotated data, many techniques have been used, and transfer learning has stood out indisputably as the most dominant paradigm. Beyond transfer learning, several other learning paradigms including unsupervised image synthesis, self-supervised and semi-supervised learning, have demonstrated great potential in performance enhancement given limited annotated data. We will introduce these learning paradigms' applications in medica image classification in the following subsections.

### 3.1.1. Supervised classification

Starting from AlexNet (Krizhevsky et al., 2012), a variety of end-to-end models with increasingly deeper networks and larger representation compacity have been developed for image classification, such as VGG (Simonyan and Zisserman, 2014), GoogleLeNet (Szegedy et al., 2015), and ResNet (He et al., 2016), and DenseNet (Huang et al., 2017). These models have yielded superior results, making deep learning mainstream not only in developing high-performing CADx schemes but also in other subfields of medical image processing.

Nonetheless, the performance of deep learning models highly depends on the size of training dataset and the quality of image annotations. In many medical image analysis tasks especially 3D scenarios, it can be challenging to establish a sufficiently large and high-quality training dataset because of difficulties in data acquisition and annotation (Tajbakhsh et al., 2016; Chen et al., 2019a). The supervised transfer learning technique (Tajbakhsh et al., 2016; Donahue et al., 2014) has been routinely used to tackle the insufficient training data problem and improve model's performance, where standard architectures like ResNet (He et al., 2016) are first pre-trained in the source domain with a large amount of natural images (e.g., ImageNet (Deng et al., 2009)) or medical images, and then the pre-trained models are transferred to the target domain and fine-tuned using much fewer training examples. Tajbakhsh et al. (2016) showed that pre-trained CNNs with adequate fine-tuning performed at least as well as CNNs trained from scratch. Indeed, transfer learning has become a cornerstone for image classification tasks (de Bruijne, 2016) across a variety of modalities, including CT (Shin et al., 2016), MRI (Yuan et al., 2019), mammography (Huynh et al., 2016), X-ray (Minaee et al., 2020), etc.

Within the paradigm of supervised classification, different types of attention modules have been used for performance boost and better model interpretability (Zhou et al., 2019b). Guan et al. (2018) introduced an attention-guided CNN, which is based on ResNet-50 (He et al., 2016). The attention heatmaps from the global X-ray image were used to suppress large irrelevant areas and highlight local regions that contain discriminative cues for the thorax disease. The proposed model effectively fused the global and local information and achieved a good classification performance. In another study, Schlemper et al. (2019) incorporated attention modules to a variant network of VGG (Baumgartner et al., 2017) and U-Net (Ronneberger et al., 2015) for 2D fetal ultrasound image plane classification and 3D CT pancreas segmentation, respectively. Each attention module was trained to focus on a subset of local structures in input images, and these local structures contain salient features useful to the target task.

### 3.1.2. Unsupervised methods

### I. Unsupervised image synthesis

Classic data augmentation (e.g., rotation, scale, flip, translation, etc.) is simple but effective in creating more training instances to achieve better performance (Krizhevsky et al., 2012). However, it cannot bring much new information to the existing training examples. Given the advantage of learning hidden data distribution and generating realistic images, GANs have been used as a more complicated approach for data augmentation in the medical domain.

Frid-Adar et al. (2018b) exploited DCGAN for synthesizing high-quality examples to improve liver lesion classification on a limited dataset. The dataset only has 182 liver lesions including cysts, metastases, and hemangiomas. Since training GAN typically needs a large number of examples, the authors applied classic data augmentation (e.g., rotation, flip, translation, scale) to create nearly 90,000 examples. The GAN-based synthetic data augmentation significantly improved the classification performance, with the sensitivity and specificity increased from 78.6% and 88.4% to 85.7% and 92.4% respectively. In their later work (Frid-Adar et al., 2018a), the authors further extended lesion synthesis from the unconditional setting (DCGAN) to a conditional setting (ACGAN). The generator of ACGAN was conditioned on the side information (lesion classes), and the discriminator predicted lesion classes besides synthesizing new examples. However, it was found that ACGAN-based synthetic augmentation delivered a weaker classification performance than its unconditional counterpart.

To alleviate data scarcity and especially the lack of positive cancer cases, Wu et al. (2018a) adopted a conditional structure (cGAN) to generate realistic lesions for mammogram classification. Traditional data augmentation was also used to create enough examples for training GAN. The generator, conditioned with malignant/non-malignant labels, can control the process of generating a specific type of lesions. For each non-malignant patch image, a malignant lesion was synthesized onto it using a segmentation mask of another malignant lesion; for each malignant image, its lesion was removed, and a non-malignant patch was synthesized. Although the GAN-based augmentation achieved better classification performance than traditional data augmentation, the improvement was relatively small, less than 1%.

## II. Self-supervised learning based classification

Recent self-supervised learning approaches have shown great potential in improving performance of medical tasks lacking sufficient annotations (Bai et al., 2019; Tao et al., 2020; Li et al., 2020a; Shorfuzzaman and Hossain, 2021; Zhang et al., 2020a). This method is suitable to the scenario where large amounts of medical images are available, but only a small percentage are labeled. Accordingly, the model optimization is divided into two steps, namely, self-supervised pre-training and supervised fine-tuning. The model is initially optimized using unlabeled images to effectively learn good features that are representative of the image semantics (Azizi et al., 2021). The pre-trained models from self-supervision are followed by supervised fine-tuning to achieve faster and better performance in subsequent classification tasks (Chen et al., 2020c). In practice, self-supervision can be created either through pretext tasks (Misra and Maaten, 2020) or contrastive learning (Jing and Tian, 2020) as follows.

Self-supervised **pretext task** based classification utilizes common pretext tasks such as rotation prediction (Tajbakhsh et al., 2019) and Rubik's cube recovery (Zhuang et al., 2019; Zhu et al., 2020a). Chen et al. (2019b) argued that existing pretext tasks such as relative position prediction (Doersch et al., 2015) and local context prediction (Pathak et al., 2016) resulted in only marginal improvements on medical image datasets; the authors designed a new pretext task based on context restoration. This new pretext task has two steps: disordering patches in corrupted images and restoring the original images. The context restoration pre-training strategy improved the performance of medical image classification. Tajbakhsh et al. (2019) exploited three pretext tasks, namely, rotation (Gidaris et al., 2018), colorization (Larsson et al., 2017), and WGAN-based patch reconstruction, to pre-train models for classification tasks. After pre-training, models were trained using labeled examples. It was shown that pretext tasks based pre-training in the medical domain was more effective than random initializations and transfer learning (ImageNet pre-training) for diabetic retinopathy classification.

For self-supervised **contrastive** classification, Azizi et al. (2021) adopted the self-supervised learning framework SimCLR (Chen et al., 2020a) to train models (wider versions of ResNet-50 and ResNet-152) for dermatology condition classification and chest X-ray classification. They pre-trained the models by first using unlabeled natural images then with unlabeled dermatology images and chest X-rays. Feature representations were learned by maximizing agreement between positive image pairs that are either two augmented examples of the same image or multiple images from the same patient. The pre-trained models were fine-tuned using much fewer labeled dermatology images and chest X-rays. These models outperformed their counterparts pre-trained using ImageNet by 1.1% in mean AUC for chest X-ray classification and 6.7% in top-1 accuracy for dermatology condition classification. MoCo (He et al., 2020; Chen et al., 2020b) is another popular self-supervised learning framework to pre-train models for medical classification tasks, such as COVID-19 diagnosis from CT images (Chen et al., 2021a) and pleural effusion identification in chest X-rays (Sowrirajan et al., 2021). Furthermore, it has been shown that self-supervised contrastive pre-training can greatly benefit from the incorporation of domain knowledge. For example, Vu et al. (2021) harnessed patient metadata (patient number, image laterality, and study number) to construct and select positive pairs from multiple chest X-ray images for MoCo pre-training. With only 1% of the labeled data for pleural effusion classification, the proposed approach improved mean AUC by 3.4% and 14.4% compared to previous contrastive learning method (Sowrirajan et al., 2021) and ImageNet pre-training respectively.

### 3.1.3. Semi-supervised learning

Unlike self-supervised approaches that can learn useful feature representations just from unlabeled data, **semi-supervised learning** needs to integrate unlabeled data with labeled data through different ways to train models for a better performance. Madani et al. (2018a) employed GAN that was trained in a semi-supervised manner (Kingma et al., 2014) for cardiac disease classification in chest X-rays where labeled data was limited. Unlike the vanilla GAN (Goodfellow et al., 2014), this semi-supervised GAN was trained using both unlabeled and labeled data. Its discriminator was modified to predict not only the realness of input images but also image classes (normal/abnormal) for real data. When increasing the number of labeled examples, the semi-supervised GAN based classifier consistently performed better than supervised CNN. Semi-supervised GAN was also shown useful in other data-limited classification tasks, such as CT lung nodule classification (Xie et al., 2019a), and left ventricular hypertrophy classification from echocardiograms (Madani et al., 2018b). Besides the semi-supervised adversarial approach, consistency-based semi-supervised methods such as $\Pi$-Model (Laine and Aila, 2017) and Mean Teacher (Tarvainen and Valpola, 2017) have also been used to leverage unlabeled medical image data for better classification (Shang et al., 2019; Liu et al., 2020a).

Table 1. A list of recent papers related to medical image classification

| Author | Year | Application | Model | Dataset | Contributions highlights |
|---|---|---|---|---|---|
| | | | **Supervised classification** | | |
| Guan et al., 2018 | 2018 | Thorax disease classification from chest X-rays | *AG-CNN*: an attention guided CNN | ChestX-ray14 | (1) Using attention mechanism to identify discriminative regions from the global image, which were used to train the local CNN branch; (2) Fusing the local and global information for better performance. |
| Schlemper et al., 2019 | 2019 | 2D fetal ultrasound image plane classification | *AG-Sononet*: attention-gated model | Private dataset | Incorporating grid attention into Sononet (Baumgartner et al., 2017) to better exploit local information and aggregating attention vectors at different scales for final prediction. |
| | | | **Unsupervised image synthesis** | | |
| Frid-Adar et al., 2018b | 2018 | CT liver lesion classification | DCGAN | Private dataset | Using the high-quality liver ROIs synthesized by DCGAN to perform data augmentation. |
| Frid-Adar et al., 2018a | 2018 | CT liver lesion classification | ACGAN | Private dataset | Comparing GAN's augmentation performance in conditional and unconditional settings. |
| Wu et al., 2018a | 2018 | Mammogram classification | cGAN | DDSM dataset | Controlling generating a specific type of lesions using malignant/non-malignant labels. |
| | | | **Self-supervised learning based classification** | | |
| Azizi et al., 2021 | 2021 | Classification of chest X-ray images and dermatology images | *MICLe*: based on | CheXpert, and private dataset | Proposing a new contrastive learning approach based on *SimCLR* by leveraging multiple |

| | | | SimCLR (Chen et al., 2020a) | | images of each medical condition for additional self-supervised pretraining. |
|---|---|---|---|---|---|
| Vu et al., 2021 | 2021 | Classification of pleural effusion in chest X-ray images | *MedAug:* Based on MoCo (Chen et al., 2020b) | CheXpert | (1) Utilizing patient metadata to create positive pairs for contrastive learning; (2) Showing that self-supervised pre-training can perform better than ImageNet pre-training. |
| Chen et al., 2021a | 2021 | COVID-19 diagnosis from chest CT images | MoCo-based classification | DeepLesion, LIDC-IDRI, UCSD COVID-19 CT, SIRM's COVID-19 data | Using contrastive learning to pre-train an encoder on public datasets so that expressive features of non-COVID CT images can be captured, and using the pre-trained encoder for few-shot COVID-19 classification. |
| Sowrirajan et al., 2021 | 2021 | Classification of pleural effusion and tuberculosis from chest X-rays | *MoCo-CXR:* Based on MoCo (Chen et al., 2020b) | CheXpert, Shenzen dataset | Showing that MoCo-pretrained feature representations on large X-ray databases can (1) outperform ImageNet pre-training on downstream tasks with small, labeled X-rays, and (2) generalize well to an external dataset. |
| Chen et al., 2019b | 2019 | Fetal ultrasound image plane classification | A general CNN-based architecture | Private dataset | Designing a new self-supervised pretext task based on context restoration to learn high-quality features from unlabeled images. |
| Zhou et al., 2021 | 2021 | CT lung nodule false positive reduction, etc. | *Models Genesis* | LUNA 2016, etc. | Consolidating four different self-supervised schemes (non-linear, local-shuffling, inner and outer cutouts) to learn representations from different perspectives (appearance, texture, and context). |
| **Semi-supervised learning based classification** | | | | | |
| Liu et al., 2020a | 2020 | Thorax disease classification from chest X-rays, etc. | Based on *Mean Teacher* (Tarvainen and Valpola, 2017) | ChestX-ray14, etc. | Proposing *sample relation consistency* for the semi-supervised model to extract useful semantic information from the unlabeled data. |
| Xie et al., 2019a | 2019 | CT lung nodule classification | Adversarial autoencoder-based model | LIDC-IDRI, Tianchi Lung Nodule dataset | Using learnable transition layers to enable transferring representations from the reconstruction network to the classification network. |
| Madani et al., 2018a | 2018 | Cardiac abnormality classification in chest X-rays | *Semi-supervised GAN* | NIH PLCO dataset, etc. | Employing a semi-supervised GAN architecture to address the scarcity of labeled data. |

### 3.2. Segmentation

Medical image segmentation, identifying the set of pixels or voxels of lesions, organs, and other substructures from background regions, is another challenging task in medical image analysis (Litjens et al., 2017). Among all common image analysis tasks such as classification and detection, segmentation needs the strongest supervision (large amounts of high-quality annotations) (Tajbakhsh et al., 2020). Since its introduction in 2015, U-Net (Ronneberger et al., 2015) has become probably the most well-known architecture for segmenting medical images; afterwards, different variants of U-Net have been proposed to further improve the segmentation performance. From the very recent literature, we observe that the combination of U-Net and Transformers from NLP (Chen et al., 2021b) has contributed to state-of-the-art performance. In addition, a number of semi-supervised and self-supervised learning based approaches have also been proposed to alleviate the need for large annotated datasets. Accordingly, in this section we will (1) review the original U-Net and its important variants, and summarize useful performance enhancing strategies; (2) introduce the combination of U-Net and Transformers, and Mask RCNN (He et al., 2017); 3) cover self-supervised and semi-supervised approaches for segmentation. Since recent studies focus on applying Transformers to segment medical images in a supervised manner, we purposely position the introduction of Transformers-based architectures in the supervised segmentation section. However, it should be noted that such categorization does not mean Transformers-based architectures cannot be used in semi-supervised or unsupervised settings.

### 3.2.1. Supervised learning based segmenting models

### I. U-Net and its variants

In a convolutional network, the high-level coarse-grained features learned by higher layers capture semantics beneficial to the whole image classification; in contrast, the low-level fine-grained features learned by lower layers contain useful details for precise localizations (i.e., assigning a class label to each pixel) (Hariharan et al., 2015), which are important to image segmentation. U-Net is built on the fully convolutional network (Long et al., 2015), the key innovation of U-Net is the so-called skip connections between opposing convolutional layers and deconvolutional layers, which successfully concatenate features learned at different levels to improve the segmentation performance. Meanwhile, skip connections is also helpful in recovering the network's output to be of the same spatial resolution as the input. U-Net takes 2D images as input, and it generates several segmentation maps, each of which corresponds to one respective pixel class.

Based on the basic architectures, Drozdzal et al. (2016) further studied the influence of long and short skip connections in biomedical image segmentation. They concluded that adding short skip connections is important to train very deep segmentation networks. In one study, Zhou et al. (2018) claimed that the plain skip connections between U-Net's encoder and decoder subnetworks leads to fusion of semantically dissimilar feature maps; they proposed to reduce the semantic gap prior to fusing feature maps. In the proposed model UNet++, the plain skip connections were replaced by nested and dense skip connections. The suggested architecture outperformed U-Net and wide U-Net across four different medical image segmentation tasks.

Aside from redesigning the skip connections, Çiçek et al. (2016) replaced all 2D operations with their 3D counterparts to extend the 2D U-Net to 3D U-Net for volumetric segmentation with sparsely annotated images. Further, Milletari et al. (2016) proposed V-Net for 3D MRI prostate volumes segmentation. A major architecture difference between U-Net and V-Net lies in the change of the forward convolutional units (Figure 5(a)) to residual convolutional units (Figure 5(c)), so V-Net is also referred as residual U-Net. A new loss function based on Dice coefficient was proposed to deal with the imbalanced number of foreground and background voxels. To tackle the scarcity of annotated volumes, the authors augmented their training dataset with random non-linear transformations and histogram matching. Gibson et al. (2018a) proposed the Dense V-network that modified V-Net's loss function of binary segmentation to support multiorgan segmentation of abdominal CT images. Although the authors followed the V-Net architecture, they replaced its relatively shallow down-sampling network with a sequence of three dense feature stacks. The combination of densely linked layers and the shallow V-Net architecture demonstrates its importance in improving segmentation accuracy, and the proposed model yielded significantly higher Dice scores for all organs compared to multi-atlas label fusion (MALF) methods.

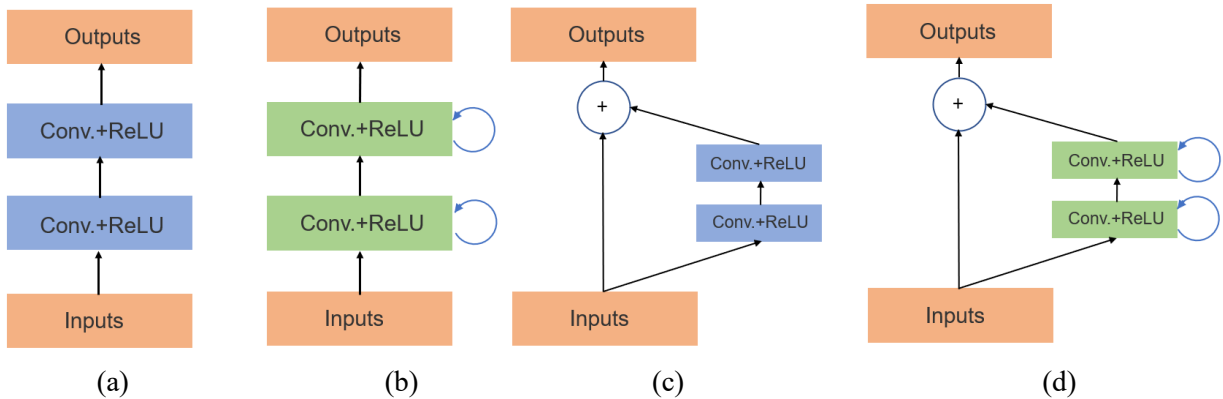

Figure 5. Units of different segmentation networks (a) forward convolutional unit (U-Net), (b) recurrent convolutional block (RCNN), (c) residual convolutional unit (residual U-Net), and (d) recurrent residual convolutional unit (R2U-Net) (Alom et al., 2018).

Alom et al. (2018) proposed to integrate the architectural advantages of recurrent convolutional neural network (RCNN) (Ming and Xiaolin, 2015) and ResNet (He et al., 2016) when designing U-Net based segmentation networks. In their first network (RU-Net), the authors replaced U-Net's forward convolutional units using RCNN's recurrent convolutional layers (RCL) (Figure 5(b)), which can help accumulate useful features to improve segmentation results. In their second network (R2U-Net), the authors further modified RCL using ResNet's residual units (Figure 5(d)), which learns a residual function by using identity mapping for shortcut connections, thus allowing for training very deep networks. Both models achieved better segmentation performance than U-Net and residual U-Net. Dense convolutional blocks (Huang et al., 2017) also demonstrated its superiority in enhancing segmentation performance on liver and tumor CT volumes (Li et al., 2018).

Besides the redesigned skip connections and modified architectures, U-Net based segmentation approaches also benefit from adversarial training (Xue et al., 2018; Zhang et al., 2020b), attention mechanisms (Jetley et al., 2018; Anderson et al., 2018; Oktay et al., 2018; Nie et al., 2018; Sinha and Dolz, 2021), and uncertainty estimation (Wang et al., 2019a; Yu et al., 2019; Baumgartner et al., 2019; Mehrtash et al., 2020). For example, Xue et al. (2018) developed an adversarial network for brain tumor segmentation, and the network has two parts: a segmentor and a critic. The segmentor is a U-Net-like network that generates segmentation maps given input images; the predicted maps and ground-truth segmentation maps are sent into the critic network. Alternatively training these two components eventually led to good segmentation results. Oktay et al. (2018) proposed incorporating attention gates (AGs) into the U-Net architecture to suppress irrelevant features from background regions and highlight important salient features that are propagated through the skip connections. Attention U-Net consistently delivered a better performance than U-Net in CT pancreas segmentation. Baumgartner et al. (2019) developed a hierarchical probabilistic model to estimate uncertainties in the segmentation of prostate MR and thoracic CT images. The authors employed variational autoencoders to infer the uncertainties or ambiguities in expert annotations, and separate latent variables were used to model segmentation variations at different resolutions.

**II. Transformers for segmentation**

Transformers are a group of encoder-decoder network architectures used for sequence-to-sequence processing in NLP (Chaudhari et al., 2021). One critical sub-module is known as *multi-head* self-attention (MSA), where multiple parallel self-attention layers are used to simultaneously generate multiple attention vectors for each input. Different from the convolution based U-Net and its variants, *Transformers* rely on the self-attention mechanisms, which possess the advantage of learning complex, long-range dependencies from input images. There exist two ways to adapt Transformers in the context of medical image segmentation: hybrid and Transformer-only. The hybrid approach combines CNNs and Transformers, while the latter approach does not involve any convolution operations.

Chen et al. (2021b) present *TransUNet*, the first Transformers-based framework for medical image segmentation. This architecture combines CNN and Transformer in a cascaded manner, where one's advantages are used to compensate for the other's limitations. As introduced previously, U-Net and its variants based on convolution operations have achieved satisfactory results. Because of skip connections, low-level/high-resolution CNN features from the encoder, which contain precise localization information, are utilized by the decoder to enable better performance. However, due to the intrinsic locality of convolutions, these models are generally weak at modeling long-range relations. On the other hand, although Transformers based on self-attention mechanisms can easily capture long-range dependencies, the authors found that using Transformer alone cannot provide satisfactory results. This is because it exclusively concentrates on learning global context but ignores learning low-level details containing important localization information. Therefore, the authors propose to combine low-level spatial information from CNN features with global context from the Transformer. As shown in Figure 6 (b), TransUNet has an encoder-decoder design with skip connections. The encoder is composed of a CNN and several Transformer layers. The input image needs to be first split into patches and tokenized. Then the CNN is used to generate feature maps for input patches. CNN features at different resolution levels are passed to the decoder though skip connections, so that spatial localization information can be retained. Next, patch embeddings and positional embeddings are applied to the sequence of feature maps. The embedded sequence is sent into a series of Transformer layers to learn global relations. Each Transformer layer consists of an MSA block (Vaswani et al., 2017; Dosovitskiy et al., 2020) and a multi-layer perceptron (MLP) block (Figure 6 (a)). The hidden feature representations produced by the last Transformer layer are reshaped and gradually upsampled by the decoder, which outputs a final segmentation mask. TransUNet demonstrates superior performance in the CT multi-organ segmentation task over other competing methods like attention U-Net.

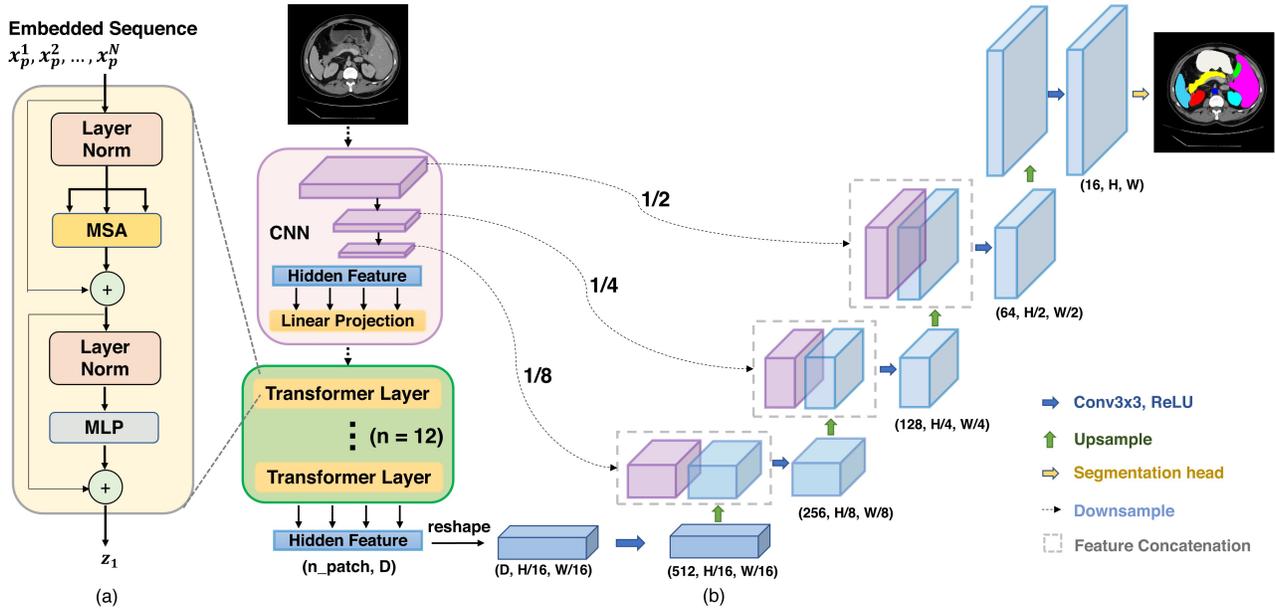

Figure 6. (a) Transformer layer; (b) the architecture of TransUNet (Chen et al., 2021b).

In another study, Zhang et al. (2021) adopt a different approach to combine CNN and Transformer. Instead of first using CNN to extract low-level features and then passing features through the Transformer layers, the proposed model *TransFuse* combines CNN and Transformer with two branches in a parallel manner. The Transformer branch consisting of several layers takes as input a sequence of embedded image patches to capture global context information. The output of the last layer is reshaped into 2D feature maps. To recover finer local details, these maps are upsampled to higher resolutions at three different scales. Correspondingly, the CNN branch uses three ResNet-based blocks to extract features from local to global at three different scales. Features with the same resolution scale from both branches are selectively fused using an independent module. The fused features can capture both the low-level spatial and high-level global context. In the end, the multi-level fused features are used to generate a final segmentation mask. TransFuse achieved good performance in prostate MRI segmentation.

In addition to 2D image segmentation, the hybrid approach is also useful to 3D scenarios. Hatamizadeh et al. (2022) propose a UNet-based architecture to perform volumetric segmentation of MRI brain tumor and CT spleen. Similar to 2D cases, 3D images are first split into volumes. Then linear embeddings and positional embeddings are applied to the sequence of input image volumes before fed to the encoder. The encoder, composed of multiple Transformer layers, extracts multi-scale global feature representations from the embedded sequence. The extracted features at different scales are all upsampled to higher resolutions and later merged with multi-scale features from the decoder via skip connections. In another study, Xie et al. (2021b) research on reducing Transformers' computational and spatial complexities in the 3D multi-organ segmentation task. To achieve this goal, they replace the original MSA module in the vanilla Transformer with the deformable self-attention module (Zhu et al., 2021a). This attention module attends over a small set of key positions instead of treating all positions equally, thus resulting in much lower complexity. Besides, their proposed architecture *CoTr,* is in the same spirit of TransUNet – a CNN generates feature maps, used as the inputs of Transformers. The difference lies in that instead of extracting only single-scale features, the CNN in CoTr extracts feature maps at multiple scales.

For the Transformer-only paradigm, Cao et al. (2021) present *Swin-Unet*, the first UNet-like pure Transformer architecture for medical image segmentation. Swin-UNet has a symmetric encoder-decoder structure without using any convolutional operations. The major components of the encoder and decoder are (1) Swin Transformer blocks (Liu et al., 2021) and (2) patching merging or expanding layers. Enabled by a shifted windowing scheme, the Swin Transformer block exhibits better modeling power as well as lower complexity in computing self-attention. Therefore, the authors use it to extract feature representations for the input sequence of image patch embeddings. The subsequent patching layer down-samples the feature representations/ maps into lower resolutions. These down-sampled maps are further passed through several other Transformer blocks and patching merging layers. Likewise, the decoder also uses Transformer blocks for feature extraction, but its patching expanding layers upsample feature maps into higher resolutions. Similar to U-Net, the upsampled feature maps are fused with the down-sampled feature maps from the encoder via skip connections. Finally, the decoder outputs pixel-level segmentation predictions. The proposed framework achieved satisfactory results on multi-organ CT and cardiac MRI segmentation tasks.

Note that, to ensure good performance and reduce training time, most of the Transformers-based segmentation models introduced so far are pre-trained on a large external dataset (e.g., ImageNet). Interestingly, it has been shown that Transformers can also produce good results without pre-training by utilizing computationally efficient self-attention modules (Wang et al., 2020a) and new training strategies to integrate high-level information and finer details (Valanarasu et al., 2021). Also, when applying Transformers-based model for 3D medical image segmentation, Hatamizadeh et al. (2022) and Xie et al. (2021b) find pre-training did not show performance improvement.

### III. Mask R-CNN for segmentation

Aside from the above UNet and Transformers-based approaches, another architecture Mask RCNN (He et al., 2017), which was originally developed for pixelwise instance segmentation, has achieved good results in medical tasks. Since it is closely related to Faster RCNN (Ren et al., 2015; Ren et al., 2017), which is a region-based CNN for object detection, details of Mask RCNN and its relations with the detection architectures will be elaborated later. To sum up in brief, Mask RCNN has (1) a region proposal network (RPN) as in Faster RCNN to produce high-quality region proposals (i.e., likely to contain objects), (2) the RoIAlign layer to preserve spatial correspondence between RoIs and their feature maps, and (3) a parallel branch for binary mask prediction in addition to bounding box prediction as in Faster RCNN. Notably, the Feature Pyramid Network (FPN) (Lin et al., 2017a) is used as the backbone of the Mask RCNN to extract multi-scale features. FPN has a bottom-up pathway and a top-down pathway to extract and merge features in a pyramidal hierarchy. The bottom-up pathway extracts feature maps from high resolution (semantically weak features) to low resolution (semantically strong), whereas the top-down pathway operates in the opposite. At each resolution, features generated by the top-down pathway are enhanced by features from the bottom-up pathway via skip connections. This design might make FPN seemingly resemble U-Net, but the major difference is that FPN makes predictions independently at all resolution scales instead of one.

Wang et al. (2019b) proposed a volumetric attention (VA) module within the Mask RCNN framework for 3D medical image segmentation. This attention module can utilize the contextual relations along the z direction of 3D CT volumes. More concretely, feature pyramids are extracted from not only the target image (3 adjacent slices with the target CT slice in the middle), but also a series of neighboring images (also 3 CT slices). Then the target and neighboring feature pyramids are concatenated at each level to form an intermediate pyramid, which carries long-range relations in the z axis. In the end, spatial attention and channel attention are applied on the intermediate and target pyramids to form the final feature pyramid for mask prediction. With this VA module incorporated, Mask RCNN could achieve lower false positives in segmentation. In another study, Zhou et al. (2019c) combined UNet++ and Mask RCNN, leading to Mask RCNN++. As mentioned earlier, UNet++ demonstrates better segmentation results using the redesigned nested and dense skip connections, so the authors use them to replace the plain skip connections of the FPN inside Mask RCNN. A large performance boost was observed using the proposed model.

### 3.2.2. Unsupervised learning based segmenting models

For medical image segmentation, to alleviate the need for a large amount of annotated training data, reserachers have adopted generative models for image synthesis to increase the number of training examples (Zhang et al., 2018b; Zhao et al., 2019a). Meanwhile, exploiting the power of unlabeled medical images seems like a much more popular choice. In contrast to the difficult and expensive high-quality annotated dataset, unlabeled medical images are often available, usually coming with a large number. Given a small medical image dataset with limited ground truth annotations and a related but unlabeled large dataset, reserachers have explored **self-supervised** and **semi-supervised** learning approches to learn useful and transferrable feature representations from the unlabled dataset, which will be discussed in this and the next section respectively.

**Self-supervised pretext tasks**: Since self-supervision via pretext tasks and contrastive learning can learn rich semantic representations from unlabeled datasets, self-supervised learning is often used to pre-train the model and enable solving downstream tasks (e.g., medical image segmentation) more accurately and efficiently when limited annotated examples are available (Taleb et al., 2020). The pretext tasks could be either designed based on application scenarios or chosen from traditional ones used in the computer vision field. For the former type, Bai et al. (2019) designed a novel pretext task by predicting anatomical positions for cardiac MR image segmentation. The self-learnt features via the pretext task were transferred to tackle a more challenging task, accurate ventricles segmentation. The proposed method achieved much higher segmentation accuracy than the standard U-Net trained from scratch, especially when only limited annotations were available.

For the latter type, Taleb et al. (2020) extended performing pretext tasks from 2D to 3D scenarios, and they investigated the effectiveness of several pretext tasks (e.g., rotation prediction, jigsaw puzzles, relative patch location) in 3D medical image segmentation. For brain tumor segmentation, they adopted the U-Net architecture, and the pretext tasks were performed on a large unlabeled dataset (about 22,000 MRI scans) to pre-train the models; then the learned feature representations were fine-tuned on a much smaller labeled dataset (285 MRI scans). The 3D pretext tasks performed better than their 2D counterparts; more importantly, the proposed methods sometimes outperformed supervised pre-training, suggesting a good generalization ability of the self-learnt features.

The performance of self-supervised pre-training could also be improved by adding other types of information. Hu et al. (2020) implemented a context encoder (Pathak et al., 2016) performing semantic inpainting as the pretext task, and they incorporated DICOM metadata from ultrasound images as weak labels to boost the quality of pre-trained features toward facilitating two different segmentation tasks.

**Self-supervised contrastive learning** based approaches: For this method, early studies such as the work by Jamaludin et al. (2017) adopted the original contrastive loss (Chopra et al., 2005b) to learn useful feature representations. In recent three years, with a surge of interest in self-supervised contrastive learning, contrastive loss has evolved from the original version to more powerful ones (Oord et al., 2018) for learning expressive feature representations from unlabeled datasets. Chaitanya et al. (2020) claimed although the contrastive loss in Chen et al. (2020a) is suitable for learning image-level (global) feature representations, it does not guarantee learning distinctive local representations that are important for per-pixel segmentation. They proposed a local contrastive loss to capture local features that can provide complementary information to boost the segmentation performance. Meanwhile, to the best of our knowledge, when computing the global contrastive loss, these authors are the first to utilize the domain knowledge that there is structural similarity in volumetric medical images (e.g., CT and MRI). In MR image segmentation with low annotations, the proposed method substantially outperformed other semi-supervised and self-supervised methods. In addition, it was shown that the proposed method could further benefit from data augmentation techniques like Mixup (Zhang et al., 2018a).

### 3.2.3. Semi- supervised learning based segmenting models

**Semi-supervised consistency regularization**: The mean teacher model is commonly used. Based on the mean teacher framework, Yu et al. (2019) introduced uncertainty estimation (Kendall and Gal, 2017) for better segmentation of 3D left atrium from MR images. They argued that on an unlabeled dataset, the output of the teacher model can be noisy and unreliable; therefore, besides generating target outputs, the teacher model was modified to estimate these outputs' uncertainty. The uncertainty-aware teacher model can produce more reliable guidance for the student model, and the student model could in turn improve the teacher model. The mean teacher model can also be improved by the transformation-consistent strategy (Li et al., 2020b). In one study, Wang et al. (2020b) proposed a semi-supervised framework to segment COVID-19 pneumonia lesions from CT scans with noisy labels. Their framework is also based on the Mean Teacher model; instead of updating the teacher model with a predefined value, they adaptively updated the the teacher model using a dynamic threshold for the student model's segmentation loss. Similarly, the student model was also adaptively updated by the teacher model. To simultaneously deal with noisy labels and the foreground-background imbalance, the authors developed a generalized version of the Dice loss. The authors designed the segmentation network in the same spirit of U-Net but made several changes in terms of new skip connections (Pang et al., 2019), multi-scale feature representation (Chen et al., 2018a), etc. In the end, the segmentation network with the Dice loss were combined with the mean teacher framework. The proposed method demonstrated high robustness to label noise and achieved better performance for pneumonia lesion segmentation than other state-of-the-art methods.

**Semi-supervised pseudo labeling**: Fan et al. (2020) presented a semi-supervised framework (Semi-InfNet) to tackle the lack of high-quality labeled data in COVID-19 lung infection segmentation from CT images. To generate pseudo labels for the unlabeled images, they first used 50 labeled CT images to train their model, which produced pseudo labels for a small amount of unlabeled images. Then the newly pseudo-labeled examples were included in the original labeled training dataset to re-train the model to generate pseudo labels for another batch of unlabled images. This process was repeated until 1600 unlabeled CT images all got pseudo-labeled. Both the labeled and pseudo-labeled examples were used to train Semi-InfNet, and its performance surpassed other cutting-edge segmentation models such as UNet++ by a large margin. Aside from the semi-supervised learning strategy, there are three critical components in the model responsible for the good performance: parallel partial decoder (PPD) (Wu et al., 2019a), reverse attention (RA) (Chen et al., 2018b), and edge attention (Zhang et al., 2019). PPD can aggregate high-level features of the input image and generate a global map indicating the rough location of lung infection regions; EA module uses low-level features to model boundary details, and RA module further refines the rough estimation into an accurate segmentation map.

**Semi-supervised generative models**: As one of the earliest works that extended generative models to semi-supervised segmentation task, Sedai et al. (2017) utilized two VAEs to segment optic cup from retinal fundus images. The first VAE was employed to learn feature embeddings from a large number of unlabeled images by performing image reconstruction; the second VAE, trained on a smaller number of labeled images, mapped input images to segmentation masks. In other words, the authors used the first VAE to perform an auxiliary task (image reconstruction) on unlabeled data, which can help the second VAE to better achieve the target objective (image segmentation) using labled data. To leverage the feature embeddings learned by the first VAE, the second VAE simultaneously reconstructed segmentation masks and latent representations of the first VAE. The utilization of additional information from unlabeled images improved segmentation accuracy. In another study, Chen et al. (2019c) also adopted a similar idea of introducing an auxiliary task on unlabeled data to facilitate performing image segmentation with limited labeled data. Specifically, the authors proposed a semi-supervised segmentation framework consisting of a UNet-like network for segmentation (target objective) and an autoencoder for reconstruction (auxiliary task). Unlike the previous study that trained two

VAEs separately, the segmentation network and reconstruction network in this framework share the same encoder. Another difference lies in that the foreground and background parts of the input image were reconstructed/generated separately, and the respective segmentation labels were obtained via an attention mechanism. This semi-supervised segmentation framework outperformed its counterparts (e.g., fully supervised CNNs) in different labeled/unlabeled data splits.

In addition to the aforementioned approaches, researchers have also explored incorporating domain-specific prior knowledge to tailor the semi-supervised frameworks for a better segmentation performance. The prior knowledge varies a lot, such as the anatomical prior (He et al., 2019), atlas prior (Zheng et al., 2019), topological prior (Clough et al., 2020), semantic constraint (Ganaye et al., 2018), and shape constraint (Li et al., 2020c) to name a few.

Table 2. A list of recent papers related to medical image segmentation

| Author | Year | Application | Model | Dataset | Contributions highlights |
|---|---|---|---|---|---|
| **U-Net and its variants** | | | | | |
| Milletari et al., 2016 | 2016 | MRI prostate volumes segmentation | *V-Net (Residual U-Net)* | PROMISE 2012 dataset | (1) Incorporation of residual learning; (2) A new loss function based on Dice coefficient to deal with class imbalance; (3) Data augmentation by applying random non-linear transformations and histogram matching. |
| Zhou et al., 2018 | 2018 | Segmentation of (1) CT lung nodules, (2) microscopic cell nuclei, (3) CT liver, and (4) colon polyps | *UNet++* | LIDC-IDRI, Data Science Bowl 2018, MICCAI 2018 LiTS, ASU-Mayo | (1) Proposing nested and dense skip connections to reduce the semantic gap before fusing feature maps; (2) Using deep supervision to enable accurate and fast segmentation. |
| Gibson et al., 2018a | 2018 | Segmentation of CT pancreas, gastrointestinal organs, and surrounding organs | *Dense V-Net* | NIH Pancreas-CT dataset and BTCV challenge dataset | (1) A new loss function extends binary segmentation to multiorgan segmentation; (2) Integrating densely linked layers into the shallow V-Net architecture. |
| Alom et al., 2018 | 2018 | Segmentation of (1) retina blood vessels, (2) skin cancer lesions, and (3) lung | *RU-Net* and *R2U-Net* | (1) DRIVE, STARE, CHASH_DB1; (2) ISIC 2017 Challenge; (3) Data Science Bowl 2017 | (1) Replacing U-Net's forward convolutional units using RCNN's recurrent convolutional layers to accumulate useful features; (2) Incorporating residual learning to train very deep networks. |
| Oktay et al., 2018 | 2018 | Multi-class CT segmentation of pancreas, spleen, and kidney | *Attention U-Net* | NIH Pancreas-CT dataset and a private dataset | (1) Incorporating attention gates into the U-Net architecture to learn important salient features and suppress irrelevant features; (2) Image grid-based gating improves attention to local regions. |
| Xue et al., 2018 | 2018 | Brain tumor segmentation from MRI volumes | *SegAN*: adversarial network with a segmentor and a critic | MICCAI BRATS datasets in 2013 and 2015 | (1) Using adversarial learning for segmentation; (2) Proposing a multi-scale $L_1$ loss function to facilitate learning local and global features. |
| Zhang et al., 2018b | 2018 | Segmentation of multi-modal cardiovascular images (CT and MRI) | Modified GAN and a U-Net based segmentor | Private dataset | (1) Training GAN by adding a cycle-consistency loss and a shape consistency loss, making the segmentor and the generator benefit from each other; (2) Updating the generator in an online manner. |
| Zhao et al., 2019a | 2019 | Segmentation of brain MRI scans | U-Net based networks and a *SD-Net* (Roy et al., 2017) based architecture | 8 publicly available MR datasets (e.g., ADNI, OASIS, etc.) | Novel data augmentation (i.e., learning complex spatial and appearance transformations to synthesize additional labeled images based on limited labeled examples). |
| Baumgartner et al., 2019 | 2019 | Segmentation of prostate MR and thoracic CT images | *PHiSeg:* a probabilistic U-Net architecture | LIDC-IDRI and a private dataset | (1) Applying conditional VAE for inference in the U-Net architecture; (2) Using a separate latent variable to control segmentation at each resolution level to hierarchically generate final segmentations. |
| **Transformers for segmentation** | | | | | |
| Chen et al., 2021b | 2021 | Segmentation of (1) CT abdominal organs, (2) MRI cardiac structures | *TransUNet:* a hybrid cascaded CNN-Transformer architecture | Synapse multi-organ segmentation dataset, ACDC challenge | (1) Combining the advantages of CNN features (low-level spatial information) and the Transformer (modeling long-range dependencies/ high-level semantics); (2) To enable precise localization, self-attentive features from Transformer layers were combined with high-resolution CNN features via skip connections. |
| Zhang et al., 2021 | 2021 | MRI prostate segmentation | *TransFuse:* CNN and Transformer in parallel | Medical segmentation decathlon | Combining CNN and Transformer with two branches in a parallel manner and proposing the BiFusion module to fuse features from the two branches. |
| Hatamizadeh et al., 2022 | 2022 | MRI brain tumor segmentation and CT spleen segmentation | *UNETR:* UNet-based architecture | Medical segmentation decathlon | (1) Directly utilizing volumetric data for 3D segmentation; (2) The Transformer was used as the main encoder. |
| Xie et al., 2021b | 2021 | CT abdominal multi-organ segmentation | *CoTr:* an encoder-decoder structure | Synapse multi-organ segmentation dataset | (1) Multiple-scale feature maps generated by a CNN were used as the inputs of Transformers; (2) Replacing the original MSA module in the vanilla Transformer with the deformable self-attention module to reduce computational and spatial complexities. |
| Cao et al., 2021 | 2021 | Segmentation of (1) CT abdominal organs, (2) MRI cardiac structures | *Swin-Unet:* a symmetric Transformer-based design | Synapse multi-organ segmentation dataset, ACDC challenge | (1) The first Transformer-only architecture for medical image segmentation without any convolutional operations; (2) Using the Swin Transformer blocks (Liu et al., 2021) for better modeling power and lower complexity in computing self-attention. |
| Valanarasu et al., 2021 | 2021 | Segmentation of ultrasound brain anatomy and microscopic gland | *MedT:* A hybrid CNN-Transformer architecture | Private dataset, MoNuSeg, etc. | (1) Proposing positive-sensitive attention gates that enable good segmentation performance even on smaller datasets; (2) Using the entire image and image patches to train a shallow global branch and a deep local branch respectively for better performance. |
| **Mask R-CNN for segmentation** | | | | | |
| Wang et al., 2019b | 2019 | Segmentation of liver tumor on CT images | Mask RCNN with volumetric attention | LiTS challenge | Proposing a volumetric attention module to utilize the contextual relations along the z direction of 3D CT volumes. |
| Zhou et al., 2019c | 2019 | Segmentation of MRI brain tumor, CT liver, and CT lung nodules | Mask RCNN with UNet++ design | BraTS 2013, LiTS challenge, LIDC-IDRI, etc. | Using the redesigned nested and dense skip connections of UNet++ to replace the plain skip connections of the FPN inside Mask RCNN for better performance. |
| **Semi-supervised segmentation** | | | | | |
| Yu et al., 2019 | 2019 | Segmentation of left atrium from 3D MR scans | *UA-MT:* Mean Teacher framework with V-Net as backbone | 2018 Atrial Challenge dataset | Enforcing the teacher model to provide more reliable guidance to the student model via uncertainty estimation, where the estimated uncertainty was used to filter out highly uncertain predictions. |
| Li et al., 2020b | 2020 | Segmentation of (1) skin lesions, (2) fundus optic disks, and (3) CT liver | *TCSM_v2:* Mean Teacher framework with U-Net- | Datasets of (1) ISIC 2017, (2) REFUGE, (3) LiTS | Imposing transformation-consistent regularizations to unlabeled images to enhance the network's generalization capacity. |
| | | | like network as backbone | | |
| Wang et al., 2020b | 2020 | Segmentation of COVID-19 pneumonia lesions from CT scans | *COPLE-Net*: Mean Teacher framework with U-Net-like network as backbone | Private dataset | (1) Adaptively updating the teacher model and the student model; (2) Developing a generalized Dice loss to deal with noisy labels and foreground-background imbalance; (3) Using new skip connections and multi-scale feature representation. |
| Fan et al., 2020 | 2020 | Segmentation of COVID-19 lung infection from CT images | *Semi-InfNet*: Inf-Net trained in a semi-supervised manner | 2 publicly available CT datasets of COVID-19 | (1) Iterative generation of pseudo labels for unlabeled images; (2) Using the parallel partial decoder to generate a rough infection map, and reverse and edge attention modules to refine the segmentation map. (3) Multi-scale training strategy (Wu et al., 2019b). |
| **Self-supervised segmentation** | | | | | |
| Bai et al., 2019 | 2019 | Cardiac MR image segmentation | Self-supervised U-Net | UK Biobank (UKB) | (1) Pre-training the network using a new pretext tasks (i.e., predicting anatomical positions) where meaningful features were learned via self-supervision; (2) Comparing three different ways for supervised fine-tuning. |
| Taleb et al., 2020 | 2020 | Segmentation of (1) brain tumor from MRI and (2) pancreas tumor from CT | Self-supervised 3D U-Net | (1) UKB and BraTS 2018, (2) part of medical decathlon benchmarks | (1) Extending traditional 2D pretext tasks to 3D, utilizing the 3D spatial context for better self-supervision; (2) A comprehensive comparison of the performance of five different pretext tasks. |
| Hu et al., 2020 | 2020 | Segmentation of (1) thyroid nodule and (2) liver/ kidney from ultrasound images | Self-supervised U-Net with VGG16 or ResNet50 as backbone | (1) DDTI ultrasound dataset and (2) a private dataset | Incorporating DICOM metadata from ultrasound images as weak labels to improve the quality of pre-trained features from the pretext task. |
| Chaitanya et al., 2020 | 2020 | Segmentation of MRI cardiac structures and prostate regions | U-Net based encoder and decoder architecture | (1) MICCAI 2017 ACDC, (2) Medical Segmentation Decathlon, (3) STACOM 2017 | (1) Proposing a local contrastive loss; (2) Incorporation of domain knowledge (structural similarity in volumetric) in contrastive loss calculation; (3) A comprehensive comparison of a variety of pre-training techniques, such as self-supervised contrastive and pretext task pre-training, etc. |
| Sedai et al., 2017 | 2017 | Optic cup segmentation from retinal fundus images | Two VAE-based models | Private dataset | Forcing the VAE to reconstruct not only segmentation masks but also latent representations so that useful information learned from unlabeled images can be leveraged. |
| Chen et al., 2019c | 2019 | Brain tumor and white matter hyperintensities segmentation from MRI scans | UNet-like network and autoencoder | BraTS18, WMH17 Challenge | Utilizing an attention mechanism to create separate segmentation labels for foreground and background areas of the input image so that the auxiliary reconstruction task and the target segmentation task can be better linked. |

### 3.3. Detection

A natural image may contain objects belonging to different categories, and each object category may contain several instances. In the computer vision field, object detection algorithms are applied to detect and identify if any instance(s) from certain object categories are present in the image (Sermanet et al., 2014; Girshick et al., 2014; Russakovsky et al., 2015). Previous works (Shen et al., 2017; Litjens et al., 2017) have reviewed the successful applications of the frameworks before 2015, such as OverFeat (Sermanet et al., 2014; Ciompi et al., 2015), RCNN (Girshick et al., 2014), and fully convolutional networks (FCN) based models (Long et al., 2015; Dou et al., 2016; Wolterink et al., 2016). As a comparison, we aim at summarizing applications of more recent object detection frameworks (since 2015), such as Faster RCNN (Ren et al., 2015), YOLO (Redmon et al., 2016), and RetinaNet (Lin et al., 2017b). In this section, we will first briefly review several recent milestone detection frameworks, including one-stage and two-stage detectors. It should be noted that, since these detection frameworks are often used in supervised and semi-supervised settings, we introduce them under these learning paradigms. Then we will cover these frameworks' applications in specific type of lesion detection and universal lesion detection. In the end, we will introduce unsupervised lesion detection based on GANs and VAEs.

#### 3.3.1. Supervised and semi-supervised lesion detection

**I. Overview of the detection frameworks**

RCNN framework (Girshick et al., 2014) is a multi-stage pipeline. Despite its impressive results in object detection, RCNN has some drawbacks namely, the multistage pipeline makes training slow and difficult to optimize; separately extracting features for each region proposal makes training expensive in disk space and time, and it also slows down testing (Girshick, 2015). These drawbacks have inspired several recent milestone detectors, and they can be categorized into two groups (Liu et al., 2020b): (1) two-stage detection frameworks (Girshick, 2015; Ren et al., 2015; Ren et al., 2017; Dai et al., 2016), which include a separate module to generate region proposals before bounding box recognition (predicting class probabilities and bounding box coordinates); (2) one-stage detection frameworks (Redmon et al., 2016; Redmon and Farhadi, 2017; Liu et al., 2016; Lin et al., 2017b; Law and Deng, 2020; Duan et al., 2019) which predict bounding boxes in a unified manner without separating the process of generating region proposals. In an image, region proposals are a collection of potential regions or candidate bounding boxes that are likely to contain an object (Liu et al., 2020b).

**Two-stage detectors:** Unlike RCNN, the Fast RCNN framework (Girshick, 2015) is an end-to-end detection pipeline employing a multi-task loss to jointly classify region proposals and regress bounding boxes. Region proposals in Fast RCNN are generated on a shared convolutional feature map rather than the original image to speed up computation. Then a Region of Interest pooling layer was applied to warp all the region proposals into the same size. The adjustments resulted in a better and faster detection performance but the speed of Fast RCNN is still bottlenecked by the inefficient process of computing region proposals. In the Faster RCNN framework (Ren et al., 2015; Ren et al., 2017), a Region Proposal Network (RPN) replaced the selective search method to produce high-quality region proposals from anchor boxes efficiently. Anchor boxes are a set of pre-determined candidate boxes of different sizes and aspect ratios to capture objects of specific classes (Ren et al., 2015). Since that time, anchor boxes have played a dominant role in top-ranked detection frameworks. Mask RCNN (He et al., 2017) is closely related to Faster RCNN but it was originally designed for pixelwise object instance segmentation. Mask RCNN also has a RPN to propose candidate object bounding boxes; this new framework extends Faster RCNN by adding an extra branch that outputs a binary object mask to the existing branch of predicting classes and bounding box offsets. Mask RCNN uses a Feature Pyramid Network (FPN) (Lin et al., 2017a) as its backbone to extract features at various resolution scales. Besides instance segmentation, Mask RCNN can be used for object detection, achieving excellent accuracy and speed.

**One-stage detectors** Redmon et al. (2016) proposed a single-stage framework YOLO; instead of using a separate network to generate region proposals, they treated object detection as a simple regression problem. A single network was used to directly predict object classes and bounding box coordinates. YOLO also differs from region proposal based frameworks (e.g., Faster CNN) in that it learns features globally from the entire image rather than from local regions. Despite being faster and simpler, YOLO has more localization errors and lower detection accuracy than Faster RCNN. Later the authors proposed YOLOv2 and YOLO9000 (Redmon and Farhadi, 2017) to improve the performance by integrating different techniques, including batch normalization, using good anchor boxes, fine-grained features, multi-scale training, etc. Lin et al. (2017b) identified that the central cause for the lagging performance of one-stage detectors is the imbalance between foreground and background classes (i.e., the training process was dominated by vast numbers of easy examples from the background). To deal with the class imbalance problem, they proposed a new *focal loss* that can weaken the influence of easy examples and enhance the contribution of hard examples. The proposed framework (RetinaNet) demonstrated higher detection accuracy than state-of-the-art two-stage detectors at that time. Law and Deng (2020) proposed CornerNet and pointed out that the prevalent use of anchor boxes in object detection frameworks, especially one-stage detectors, causes issues such as the extreme imbalance between positive and negative examples, slow training, introducing extra hyperparameters, etc. Instead of designing a set of anchor boxes to detect bounding boxes, the authors formulated bounding boxes detection as detecting a pair of key-points (top-left and bottom-right corners) (Newell et al., 2017; Tychsen-Smith and Petersson, 2017). Nonetheless, CornerNet generates a large number of incorrect bounding boxes since it cannot fully utilize the recognizable information inside the cropped regions (Duan et al., 2019). Based on CornerNet, Duan et al. (2019) proposed CenterNet that detects each object using a triplet of key-points, including a pair corners and one center key-point. Unlike CornerNet, CenterNet can extract more recognizable visual patterns within each proposed region, thus effectively suppress inaccurate bounding boxes (Duan et al., 2019).

**II. Specific-type medical object (e.g., lesion) detection**

Common computer-aided detection (CADe) tasks include detecting lung nodules (Gu et al., 2018; Xie et al., 2019b), breast masses (Akselrod-Ballin et al., 2017; Ribli et al., 2018), lymph nodes (Zhu et al., 2020b), sclerosis lesions (Nair et al., 2020), etc. The general detection frameworks, originally designed for general object detection in natural images, cannot guarantee satisfactory performance for lesion detection in medical images for two main reasons: (1) lesions can be extremely small in size compared to natural objects; (2) lesions and non-lesions often have similar appearances (e.g. texture and intensity) (Tao et al., 2019; Tang et al., 2019). To deliver good detection performance in the medical domain, these frameworks need to be adjusted through different methods, such as incorporating domain-specific characteristics, uncertainty estimation, or semi-supervised learning strategy, which are presented as follows.

Incorporating domain-specific characteristics has been a popular choice in both the radiology and histopathology domains. In the radiology domain, the intrinsic 3D spatial context information among volumetric images (e.g. CT scans) has been utilized in many studies (Roth et al., 2016; Dou et al., 2017; Yan et al., 2018a; Liao et al., 2019). For example, in the task of pulmonary nodule detection, Ding et al. (2017) argued that the original Faster RCNN (Ren et al., 2015) with the VGG-16 network (Liu and Deng, 2015) as its backbone cannot capture representative features of small pulmonary nodules; they introduced a deconvolutional layer at the end of Faster RCNN to recover fine-grained features that are important in detecting small objects. On the deconvolutional feature map, an FPN was applied to propose candidate regions of nodules from 2D axial slices. To reduce the false positive rate, the authors proposed to make the classification network see the full range of contexts of the nodule candidates. Instead of using 2D CNN, they chose a 3D CNN to exploit the 3D context of candidate regions so that more distinctive features can be captured for nodule recognition. The proposed method ranked the 1$^{st}$ place in nodule detection on the LUNA16 benchmark dataset (Setio et al., 2017). Zhu et al. (2018) also considered the 3D nature of lung CT images and designed a 3D Faster RCNN for nodule detection. To efficiently learn nodule features, the 3D faster RCNN had the U-Net-like structure (Ronneberger et al., 2015) and was built with compact dual path blocks (Chen et al., 2017). It should be noted that despite the effectiveness in boosting detection performance, 3D CNN has downsides as compared to 2D CNN, including consuming more computational resources and requiring more efforts to acquire 3D bounding box annotations (Yan et al., 2018a; Tao et al., 2019). In a recent study, Mei et al. (2021) established a large dataset (PN9) with more than 40, 000 annotated lung nodules to train 3D CNN-based models. The authors improved the model's ability to detect both large and small lung nodules by utilizing correlations that exist among multiple consecutive CT slices. Given a slice group, a non-local operation based module (Wang et al., 2018) was employed to seize long-range dependencies of different positions and different channels in the feature map. Furthermore, since each shallow ResNet block can generate feature maps on the same scale that carry useful spatial information, the authors reduced false positive nodule candidates by merging multi-scale features produced by 3 different blocks.

In the histopathology domain, Rijthoven et al. (2018) presented a modified version of YOLOv2 (Redmon and Farhadi, 2017) for lymphocytes detection in whole-slide images (WSI). Based on the prior knowledge of lymphocytes (e.g., average size, no overlaps), the authors simplified the original YOLO network with 23 layers by keeping only a few layers. With the prior knowledge that brown areas without lymphocytes in the WSI contain many hard negative samples, the authors also designed a sampling strategy to enforce the detection model to focus on these hard negative examples during training. The proposed method improved F1-score by 3% with a speed-up of 4.3X. In their later work, Swiderska-Chadaj et al. (2019) modified the YOLO architecture to further detect lymphocytes in a more diversified WSI dataset of breast, prostate, and colon cancer; however, it did not perform as well as the U-Net based detection architecture, which first classified each pixel and then produced detection results using post-processing techniques. The modified YOLO architecture was also shown the least robust to different staining.

Recently, semi-supervised methods have been used to improve the performance of medical object detection (Gao et al., 2020; Qi et al., 2020). For example, Wang et al. (2020c) developed a generalized version of the original focal loss (Lin et al., 2017b) to deal with soft labels in computing semi-supervised loss function. They modified the semi-supervised approach MixMatch (Berthelot et al., 2019) from two aspects to make it suitable for 3D medical image detection. An FPN was first applied on unlabeled CT images (without lesion annotations) to generate pseudo-labeled object instances. Then the pseudo-labeled examples were mixed with examples having ground truth annotations through Mixup augmentation. However, the original Mixup augmentation (Zhang et al., 2018a) was designed for classification tasks where labels are image classes; the authors adapted this augmentation technique to the lesion detection task with annotations in the form of bounding boxes. The semi-supervised approach demonstrated a significant performance gain over supervised learning baselines in pulmonary nodule detection.

In addition, uncertainty estimation is another useful technique to facilitate the detection of small objects (Ozdemir et al., 2017; Nair et al., 2020). For example, in the task of multiple sclerosis lesion detection where uncertainties mostly result from small lesions and lesion boundaries, Nair et al. (2020) explored using uncertainty estimates to improve detection performance. Specifically, four uncertainty measures were computed: a predicted variance from training data (Kendall and Gal, 2017), variance of Monte Carlo (MC) samples, a predictive entropy, and mutual information. A threshold formed by these measures was used to filter out the most uncertain lesion candidates and thus improve detection performance.

**III. Universal lesion detection**

Traditional lesion detectors have focused on a specific type of lesions but there is a rising research interest in identifying and localizing different kinds of lesions from the whole human body all at once (Yan et al., 2018a; Yan et al., 2019; Tao et al., 2019; Yan et al., 2020; Cai et al., 2020; Li et al., 2020d). DeepLesion is a large and comprehensive dataset (32K lesions) that contains a variety of lesion types such as lung nodule, liver tumor, abdominal mass, pelvic mass, etc. (Yan et al., 2018b; Yan et al., 2018c). Tang et al. (2019) proposed ULDor based on Mask RCNN for universal lesion detection. Training Mask-RCNN requires ground truth masks for lesions; however, the DeepLesion dataset does not contain such annotated masks. With the RECIST (Response Evaluation Criteria In Solid Tumors) annotations (Eisenhauer et al., 2009), the authors estimated real masks via ellipse fitting for each lesion region. In addition, hard negative examples were used to re-train the model to reduce false positives. Yan et al. (2019) further improved the performance of universal lesion detection by enforcing a multitask detector (MULAN) to jointly perform lesion detection, tagging, and segmentation. It was previously shown that combining different tasks may provide complementary information to each other and thus enhance the performance of a single task (Wu et al., 2018b; Tang et al., 2019). MULAN is modified from Mask RCNN (He et al., 2017) with three head branches. The detection branch predicts whether each proposed region is lesion and regresses bounding boxes; the tagging branch predicts 185 tags (e.g., body part, lesion type, intensity, shape, etc.) for each lesion proposal; the segmentation branch outputs a binary mask (lesion/non-lesion) for each proposed region. MULAN significantly surpassed previous lesion detection models such as ULDor (Tang et al., 2019) and 3DCE (Yan et al., 2018a). Furthermore, Yan et al. (2020) have recently shown that learning from heterogenous lesion datasets and partial labels can also boost detection performance.

In addition to the above strategies, attention mechanism is another useful way to improve lesion detection. Tao et al. (2019) trained a universal lesion detector on the DeepLesion dataset, and the attention mechanism (Wang et al., 2017; Woo et al., 2018) was introduced to incorporate 3D context and spatial information into a R-FCN based detection architecture (Dai et al., 2016). A contextual attention module outputs a vector indicating the importance of features learned from different axial CT slices, so the detection framework can adaptively aggregate features from different slices (i.e., enhancing relevant contextual features); a spatial attention module outputs a weight matrix so that discriminative regions on feature maps can be amplified, through which richer and more representative features can be well learned for small lesions. The proposed method demonstrated a significant performance improvement despite using much less fewer slices. Li et al. (2019) presented an FPN based architecture with an attention module that can incorporate clinical knowledge. In clinical practice, it is common for radiologists to inspect multiple CT windows for an accurate lesion diagnosis. The authors first employed three FPNs to generate feature maps from three frequently inspected windows; then the attention module (Woo et al., 2018) was used to reweight feature maps from different windows. The prior knowledge of lesion positions was also incorporated to further improve the performance.

We observe that, whether in the detection of specific-type of lesions or universal lesions, two-stage detectors are still quite prevalent for their high performance and robustness; however, separately generating region proposals might hinder developing streamlined CADe schemes. Several very recent studies have demonstrated that good detection performance can also be obtained by one-stage detectors (Pisov et al., 2020; Lung et al., 2021; Zhu et al., 2021b). We predict that advanced anchor-free one-stage detectors (e.g., CenterNet (Duan et al., 2019)) if adjusted properly to accommodate the uniqueness of medical images, will attract much more attention and even become a better choice than two-stage detectors for developing new CADe schemes in the long run.

### 3.3.2. Unsupervised lesion detection (non-prespecified type of lesion detection)

As mentioned in the above subsections, no matter it is specific-type or universal lesion detection, certain amounts of supervision are necessary to train one-stage or two-stage detectors. To establish the supervision, types of lesions need to be prespecified before training the detectors. Once trained, the detectors cannot detect lesions not contained in the training dataset. On the contrary, unsupervised lesion detection does not require ground-truth annotations, thus the lesion types do not need to be prespecified beforehand. The unsupervised detection has the potential to detect arbitrary type of lesions (Baur et al., 2021), but its performance is not comparable to that of fully-supervised/semi-supervised methods. Despite that, it can be used to establish a rough detection of suspicious areas and provide imaging biomarker candidates.

To avoid potential confusion, we make two following clarifications. First, methods to be introduced in this subsection originate from "unsupervised anomaly detection", since it is natural to consider lesions like brain tumors as one type of anomaly in medical images. The term "anomaly detection" will be used frequently throughout the context. Second, it should be noted that "anomaly detection" often appears with another term "anomaly segmentation" in the literature (Baur et al., 2021). This is because they are essentially two closely connected tasks – once anomalous regions are detected in an image, the segmentation map can be obtained by applying a binarization threshold to the detection map. In other words, approaches applicable to one direction are usually suitable to the other, so readers will see the term "anomaly segmentation".

The core assumption of unsupervised anomaly detection is that the underlying distribution of normal parts (e.g. healthy tissues and anatomy) in the images can be captured by unsupervised models, but abnormal parts such as tumors deviate from the normative distribution, so these anomalies can be detected. Commonly used models for estimating the normative distribution mainly stem from the concept of VAE and GAN, and the success of these unsupervised models has mostly been seen in MRI. Notably, Baur et al. (2021) review a variety of autoencoders-based anomaly segmentation methods in brain MR images. The authors conduct a thorough comparison of these models and present many interesting insights into successful applications. One important conclusion reached by this paper is that restoration-based approaches generally perform better than reconstruction-based ones when runtime is not a concern. In contrast to this comprehensive review paper, we will briefly introduce reconstruction-based approaches and narrow our focus to recent works related to restoration-based detection.

In the **reconstruction-based** pargdigm, an AE- or VAE-based model projects an image into low-dimensional latent space and then reconstructs the original image from its latent representation. Only healthy images are used for training, and the model is optimized to generate low pixel-wise reconstruction error. When unhealthy images pass through the the model, the reconstruction error is expected to be low regarding normal regions but high for anomalous areas. Uzunova et al. (2019) employed a CVAE to learn latent representations of healthy image patches. Besides the reconstruction error, they further assumed a large distance between the latent representations of healthy and unhealthy patches. Combining these two distances together, the CVAE-based model delivered resonable segmentation results on MRIs with tumors. It is worthy noting that the authors integrated local context into CVAE by utilizing the relative positions of patches as condition. The location-related condition can provide additional prior information of healthy and unhealthy tissues to improve performance.

In the **restoration-based** paradigm, the target to be restored is either (1) an optimal latent representation or (2) the healthy counterpart of the input anomalous image. Both GAN-based and VAE-based methods have been applied, but GAN is generally used during latent representation restoration for the **first** type. Although the generator of GAN can easily map latent vectors back to images, it lacks the capability to perform inverse mapping, (i.e., images to latent space), which is important in calculating anomaly score. This is a key issue tackled by many works adapting GAN for anomaly detection. As a pionerring work, Schlegl et al. (2017) proposed the so-called AnoGAN to obtain the inverse mapping, the authors first pre-trained a GAN (a generator and a discriminator) using healthy images to learn the normative distribution, and kept this model's weights fixed. Then given an input image (either normal or anomalous), gradient descent in the latent space (regarding latent variable) is performed to restore the corresponding optimal latent representation. More concretely, the optimization is guided by two combined losses, namely residual loss and discrimination loss. The residual loss, just like the previously mentioned reconstruction error, measures the pixel-wise dissimilarity between real input images and images that are generated by the generator from latent variable. Meanwhile, these two types of images are sent into the discriminator network, and one intermediate layer is used to extract features for them. The difference of intermediate feature represenations is computed, resulting in the discrimination loss. Last, after optimizing on the latent variable, the authors use both losses to calculate an anomaly score, indicating whether the input image contains anomalous regions. AnoGAN delievers good performance, but iterative optimization is time-consuming. In their follow-up work, Schlegl et al. (2019) proposed a more efficient model f-AnoGAN by introducing an additional encoder, which can perform fast inverse mapping from image space to latent space. Similar to developing AnoGAN, they first pre-trained a WGAN using healthy images and again kept the models's weights fixed. Then the generator with fixed weights was employed as the decoder of an AE without futher training, whereas this AE's encoder was trained using a combination of two loss functions as introduced in AnoGAN. Once fully trained, the encoder network can efficiently map images to latent space with one single forward pass. Slightly earlier than f-AnoGAN, Baur et al. (2018) proposed the so-called AnoVAEGAN that combines VAE and GAN for fast inverse mapping. In this framework, GAN's generator and VAE's decoder are the same network, and VAE's encoder is employed to learn the inverse mapping. Therefore, three components including encoder, decoder and discriminator need to be trained. The loss function here differs from that of AnoGAN and f-AnoGAN, but it still has the reconstruction error. Also, in contrast to these two patch-based models, AnoVAEGAN directly takes the entire MR images as input and thereby can capture and utilize global context potentially valuable to anomaly segmentation.

For the **second** type, restoring a healthy counterpart of the input image means, if the input contains abnormal regions, they are expected to be removed in the restored version, while the rest normal areas are retained. Thus, a pixel-wise dissimilarity map between the input and restored image can be acquired, and anomalies can be detected. Successful restoration typically relies on maximum a posteriori (MAP) estimation. Specifically, the posterior being maximized is composed of a normative distribution of healthy images and a data consistency term (Chen et al., 2020d). The normative distribution can be modeled through a VAE or its variants, and its training is guided by ELBO, an estimation for VAE's orginal objective function (Kingma and Welling, 2013). As for the data consistency term, it controls to what extent the restored image should resemble the input. In the task of detecting brain tumors from MR images, You et al. (2019) first employ a GMVAE to capture the distribution of lesion-free MR images, and adopt the total variation norm for data consistency regularization. Then these two elements together steer the optimization in MAP estimation so that the healthy counterpart of an anomalous input is iteratively restored. Recently in their following work, Chen et al. (2021c) claim that ELBO may not be a good approximation for VAE's original loss function. As a result, this inaccurate loss could lead to learning an inaccurate normative distribution, making gradient computation in iterative optimization deviate from the true direction. To solve this issue, the authors propose using the derivatives of local Gaussian distributions to replace the gradients of ELBO. When detecting glioblastomas and gliomas on MR images, the proposed approach demonstrates higher accuracy at low false positive rates compared to other methods. Also, different from most of previous works that depend on 2D MR slices, the authors incorporate 3D information into VAE's training to further improve performance.

Table 3. A list of recent papers related to medical image detection

| Author | Year | Application | Model | Dataset | Contributions highlights |
|---|---|---|---|---|---|
| **Specific-type medical objects detection** | | | | | |
| Ding et al., 2017 | 2017 | Lung nodules detection from CT images | Faster RCNN with changed VGG16 as backbone | LUNA16 | (1) Using deconvolutional layer to recover fine-grained features; (2) Using 3D CNN to exploit 3D spatial context information for false positives reduction. |
| Zhu et al., 2018 | 2018 | Lung nodules detection from CT images | 3D Faster RCNN with U-Net-like structure, built with dual path blocks | LIDC-IDRIs | (1) Using 3D Faster RCNN considering the 3D nature of lung CT images; (2) Utilizing the compactness (i.e., fewer parameters) of dual path networks on small dataset. |
| Wang et al., 2020c | 2020 | Lung nodules detection from CT images | 3D variant of FPN with modified residual network as backbone | LUNA16 and NLST | (1) A semi-supervised learning strategy to leverage unlabeled images in NLST; (2) Mixup augmentation for examples with pseudo labels and ground truth annotations; (3) FPN outputs multi-level features to enhance small object detection. |
| Mei et al., 2021 | 2021 | Lung nodules detection from CT images | U-shaped architecture, with 3D ResNet50 as encoder | PN9 | (1) Inserting non-local modules in residual blocks to seize long-range dependencies of different positions and different channels. (2) Using multi-scale features for false positives reduction. |
| Ma et al., 2021a | 2020 | Breast mass detection from mammograms | *CVR-RCNN*: Two-branch Faster RCNNs, with relation modules (Hu et al., 2018b) | DDSM and a private dataset | Extraction of complementary relation features on CC and MLO views of mammograms using relation modules. |
| Liu et al., 2020c | 2020 | Breast mass detection from mammograms | *BG-RCNN*: Incorporating Bipartite Graph convolutional Network (BGN) into Mask RCNN | DDSM and a private dataset | (1) Modeling relations (e.g., complementary information and visual correspondences) between CC and MLO views of mammograms using BGN; (2) Defining simple pseudo landmarks in mammograms to facilitate learning geometric relations. |

| Reference | Year | Task | Method | Dataset | Highlights |
|---|---|---|---|---|---|
| Rijthoven et al., 2018 | 2018 | Lymphocytes detection in whole-slide (WSI) histology images of breast, colon, and prostate cancer | Smaller YOLOv2 with much fewer layers | Private dataset | (1) Simplifying the original YOLO network using prior knowledge of lymphocytes (e.g., average size, no overlaps); (2) Designing a new training sampling strategy using the prior knowledge (i.e., brown areas without lymphocytes contain hard negative samples). |
| Lin et al., 2019 | 2019 | Lymph node metastasis detection from WSI histology images | Modified Fully convolutional network (FCN) based on VGG16 | Camelyon16 dataset and ISBI 2016 | (1) Utilizing FCN for fast gigapixel-level WSI analysis; (2) Proposing anchor layers for model conversion to ensure dense scanning; (3) Hard negative mining. |
| Nair et al., 2020 | 2020 | Multiple sclerosis lesion detection from MR brain images | 3D U-Net based segmentation network to obtain lesions | Private dataset | (1) Uncertainty estimation using Monte Carlo (MC) dropout; (2) Using multiple uncertainty measures to filter out uncertain predictions of lesion candidates. |
| **Universal lesion detection** | | | | | |
| Yan et al., 2018a | 2018 | Detection of lung, mediastinum, liver, soft tissue, pelvis, abdomen, kidney, and bone lesions from CT images | *3DCE*: Modified R-FCN | DeepLesion | (1) Exploiting 3D context information; (2) Leveraging pre-trained 2D backbones (VGG-16) for transfer learning. |
| Tang et al., 2019 | 2019 | Detection of various types of lesions in DeepLesion | *ULDor*: Mask RCNN with ResNet-101 as backbone | DeepLesion | (1) Pseudo mask construction using RECIST annotations; (2) Hard negative mining to learn more discriminative features for false positives reduction. |
| Yan et al., 2019 | 2019 | Detection of various types of lesions in DeepLesion | *MULAN*: Modified Mask RCNN with DenseNet-121 as backbone | DeepLesion | (1) Jointly performing three different tasks (detection, tagging, and segmentation) for better performance; (2) A new 3D feature fusion strategy. |
| Tao et al., 2019 | 2019 | Detection of various types of lesions in DeepLesion | Improved R-FCN | DeepLesion | Contextual attention module aggregates relevant context features, and spatial attention module highlights discriminative features for small objects. |
| Li et al., 2019 | 2019 | Detection of various types of lesions in DeepLesion | *MVP-Net*: a three pathway architecture with FPN as backbone | DeepLesion | Using an attention module to incorporate clinical knowledge of multi-view window inspection and position information. |
| **Unsupervised lesion detection** | | | | | |
| Baur et al., 2021 | 2021 | Segmentation/detection of brain MRI | A collection of VAE- and GAN-based models | Private data, MSLUB, MSSEG2015 | A comprehensive and in-depth investigation into the strengths and shortcomings of a variety of methods for anomaly segmentation. |
| Chen et al., 2021c | 2021 | Detection of MRI brain tumors and stroke lesions | VAE-based model | CamCAN, BRATS17, ATLAS | Proposing a more accurate approximation of VAE's original loss by replacing the gradients of ELBO with the derivatives of local Gaussian distributions. |
| Chen et al., 2020d | 2020 | MRI glioma and stroke detection | VAE-based model | CamCAN, BRATS17, ATLAS | Using autoencoding-based methods to learn a prior for healthy images and using MAP estimation to for image restoration. |
| Schlegl et al., 2017 | 2017 | Anomaly detection in optical coherence tomography (OCT) | *AnoGAN*: DCGAN-based model | Private dataset | (1) The first work using GAN for anomaly detection; (2) Proposing a new approach that iteratively maps input images back to optimal latent representations for anomaly detection. |
| Schlegl et al., 2019 | 2019 | OCT anomaly detection | WGAN-based model | Private dataset | Based on *AnoGAN*, an additional encoder was introduced to perform fast inverse mapping from image space to latent space. |
| Baur et al., 2018 | 2018 | MRI multiple sclerosis detection | *AnoVAEGAN*: a combination of VAE and GAN | Private dataset | (1) Combining VAE and GAN for fast inverse mapping; (2) The model can operate on an entire MR slice to exploit global context. |

| Uzunova et al., 2019 | 2019 | MRI brain tumor detection | CVAE-based model | BRATS15 | Utilizing location-related condition to provide additional prior information of healthy and unhealthy tissues for better performance. |

## 3.4. Registration

Registration, the process of aligning two or more images into one coordinate system with matched contents, is also an important step in many (semi-)automatic medical image analysis tasks. Image registration can be sorted into two groups: rigid and deformable (non-rigid). In rigid registration, all the image pixels uniformly experience a simple transform (e.g., rotation), while deformable registration aims to establish a non-uniform mapping between images. In recent years, there have been more applications of deep learning related to this research topic, especially for deformable image registration. Similar to the organization of the review article (Haskins et al., 2020), deep learning-based medical image registration approaches in our survey are categorized into three groups: (1) deep iterative registration; (2) supervised registration; (3) unsupervised registration. Interested readers can refer to several other excellent review papers (Fu et al., 2020; Ma et al., 2021b) for a more comprehensive set of registration methods.

### 3.4.1. Deep iterative registration

In **deep iterative registration**, deep learning models learn a metric that quantifies the similarity between a target/moving image and a reference/fixed image; then the learned similarity metric is used in conjunction with traditional optimizers to iteratively update the registration parameters of classical (i.e., non-learning-based) transformation frameworks. For example, Simonovsky et al. (2016) used a 5-layer CNN to learn a metric to evaluate the similarity between aligned 3D brain MRI T1–T2 image pairs, and then incorporated the learnt metric into a continuous optimization framework to complete deformable registration. This deep learning based metric outperformed manually defined similarity metrics such as mutual information for multimodal registration (Simonovsky et al., 2016). In essence, this work is most related to previous approach in Cheng et al. (2018) that estimates the similarity of 2D CT–MR patch pairs using an FCN pre-trained with stacked denoising autoencoder; the major difference between these two works lies in network architecture (CNN vs. FCN), application scenario (3D vs. 2D), and training strategy (from scratch vs. pre-training). For T1–T2 weighted MR images and CT–MR images, Haskins et al. (2019) claimed it is relatively easy to learn a good similarity metric because these multimodal images share large similar views or simple intensity mappings. They extended the deep similarity metric to a more challenging scenario, 3D MR–TRUS prostate image registration, where a large appearance difference exists between the two imaging modalities.

In summary, "deep similarity", which can avoid manually defining similarity metrics, is useful for establishing pixel-to-pixel and voxel-to-voxel correspondences. Deep similarity remains an important research track, and it is often mentioned interchangeably with several other terms like "metric learning" and "descriptor learning" (Ma et al., 2021b). Note that methods related to reinforcement learning can also be used to implicitly quantify image similarity, but we do not expand on this topic since reinforcement learning is beyond the scope of this review paper. Instead, more advanced deep similarity based approaches (e.g., adversarial similarity) will be reviewed in the unsupervised registration subsection.

### 3.4.2 Supervised registration

Despite the success of deep iterative registration, the process of learning a similarity metric followed by iterative optimization in classic registration frameworks is too slow for real-time registration. In comparison, some **supervised registration methods** directly predict deformation fields/transformations in just one step, bypassing the need for iterative optimization. These methods typically require ground truth warp/deformation fields, which can be synthesized/simulated Uzunova et al. (2017), manually annotated, or obtained via classical registration frameworks. For 3D deformable image registration, Sokooti et al. (2017) developed multi-scale CNNs based model to directly predict displacement vector fields (DVFs) between image pairs. To make their training dataset larger and more diversified, they first artificially generated DVFs with varying spatial frequency and amplitude, and then applied data augmentation on the generated DVFs, resulting in approximately 1 million training examples. After training, deformed images were registered in one-shot, and their method demonstrated close performance to a conventional B-spline registration.

Besides the supervision from ground truth deformation fields, image similarity metrics are sometimes incorporated to provide additional guidance for more accurate registration. Such combination is referred to as "dual supervision." In a recent study, Fan et al. (2019a) developed a dual-supervised training strategy with dual guidance for brain MR image registration. With the ground truth guidance, the difference between the ground truth field and predicted deformation field was calculated. With the image similarity guidance, the authors computed the difference between the template image and subject image that was warped using the predicted deformation field. The former guidance enabled the network to converge fast, while the latter guidance further refined training and yielded more accurate registration results.

### 3.4.3. Unsupervised registration

**Unsupervised learning** based registration has received extensive attention in recent years (Zhao et al., 2019b; Kim et al., 2019) for two major reasons: (1) It is cumbersome to obtain ground truth warp fields via conventional registration methods; (2) Types of deformations used for model training are limited, resulting in unsatisfactory performance on unseen images. As one of the early works related to unsupervised registration, Wu et al. (2016) argued that supervised learning based registration methods do not generalize well on new data; they employed a convolutional stacked autoencoder (Lee et al., 2011) to extract features from fixed and moving images to improve registration performance.

Balakrishnan et al. (2018) proposed an unsupervised registration model (VoxelMorph in Figure 7) that does not need supervised information (e.g., true registration fields or anatomical landmarks). The model has two components, including a convolutional U-Net and a spatial transformer network (STN). The authors formulated 3D MR brain volume registration as a parametric function, which was modeled using the U-Net architecture. The encoder's input is the concatenation of a moving image and a fixed image, and the decoder outputs a registration field. The spatial transformer network (Jaderberg et al., 2015) was applied to warp the moving image with the learned registration field, resulting in a reconstructed version of the fixed image. By minimizing the difference between the reconstructed image and the fixed image, VoxelMorph can update parameters for generating desired deformation fields. This unsupervised registration framework was able to operate orders of magnitude faster but achieved competitive performance to Symmetric Normalization (SyN) (Avants et al., 2008), a classic registration algorithm. In a later paper (Balakrishnan et al., 2019), the authors extended VoxelMorph to leverage auxiliary segmentation information (anatomical segmentation maps), and the extended model demonstrated an improved registration accuracy. Prior to this, several works had shown when there is no ground truth for voxel-level transformation, solely using auxiliary anatomical information can achieve accurate cross-modality registration (Hu et al., 2018c; Hu et al., 2018d). Note that the inclusion of segmentation information from corresponding anatomical structures is often referred to as "weakly supervised registration".

DLIR is another famous unsupervised registration framework (de Vos et al., 2019), which is an extension of the previous work (de Vos et al., 2017). DLIR has four stages to progressively perform image registration. The first stage is designed for affine image registration (AIR), and the rest three stages are for deformable image registration (DIR). In the AIR stage, a CNN takes as input pairs of fixed and moving images and outputs predictions for the affine transformation parameters so that affinely aligned image pairs can be obtained. In the subsequent DIR stage, these aligned image pairs are the input of a new CNN, whose output is a B-spline displacement vector as the deformation field. With this field, deformably registered image pairs can be obtained, and the registration results are further refined through the rest two DIR stages.

The unsupervised registration frameworks described above all utilize manually defined similarity metrics and certain regularization terms to design their loss functions. For instance, the loss function of

VoxelMorph consists of a similarity metric (mean squared error, cross-correlation (Avants et al., 2008)) to quantify the voxel correspondence between the warped image and the fixed image and a regularization term to control the spatial smoothness of the warped image (Balakrishnan et al., 2019). Despite the effectiveness of classical similarity measures in mono-modal registration, they receive less success than deep similarity metrics in most multi-modal cases. To this end, advanced deep similarity metrics learned under unsupervised regimes have been proposed to achieve superior results for multi-modal registration. One notable example is the adversarial similarity proposed by Fan et al. (2019b). Specifically, the authors proposed an unsupervised adversarial network, with a UNet-based generator and a CNN-based discriminator. The generator takes two input image volumes (moving image and fixed image) and outputs a deformation field, whereas the discriminator determines whether a negative pair of images (the fixed image and the moving image warped using the predicted field) are well-registered by comparing their similarity with a positive pair of images (the fixed image and a reference image). Using the feedback from the discriminator to improve itself, the generator is trained to generate as accurate deformations as possible to fool the discriminator. This unsupervised adversarial similarity network yielded promising results for mono-modal brain MRI image registration and multi-modal pelvic image registration.

Table 4. A list of recent papers related to medical image registration

| Author | Year | Application | Model | Dataset | Contributions highlights |
|---|---|---|---|---|---|
| **Supervised registration** | | | | | |
| Haskins et al., 2019 | 2019 | 3D MR–TRUS prostate image registration | CNN-based network with a skip connection | Private dataset | (1) Using the designed CNN to learn a similarity metric for rigid registration; (2) Proposing a new strategy to perform the optimization. |
| Cheng et al., 2018 | 2018 | 2D CT-MR patches registration | FCN pre-trained with stacked denoising AE | Private dataset | Learning a metric via FCN to evaluate the similarity between 2D CT-MR image patches for deformable registration. |
| Simonovsky et al., 2016 | 2016 | Registration of T1 and T2-weighted MRI scans | 5-layer CNN | ALBERTs | Learning a metric via CNN to evaluate the similarity between aligned 3D brain MRI T1–T2 image pairs for deformable registration. |
| Yang et al., 2017 | 2017 | Atlas-to-image and image-to-image registration | A deep encoder-decoder network | OASIS, IBIS 3D Autism Brain dataset | (1) Using deep nets to predict the momentum-parameterization of LDDMM; (2) A probabilistic version of the prediction network was developed to calculate uncertainties in the predicted deformations. |
| Fan et al., 2019a | 2019 | Brain MR image registration | *BIRNet:* hierarchical dual-supervised FCN | LPBA40, IBSR18, CUMC12, IXI30 | Providing coarse guidance (pre-registered ground-truth deformation field) and fine guidance (similarity metric) to refine the registration results. |
| Sokooti et al., 2017 | 2017 | 3D chest CT image registration | *RegNet:* a new CNN-based architecture | Private dataset | (1) Training the model using artificially generated DVFs without defining a similarity metric; (2) Incorporating contextual information into the network by processing input 3D image patches at at multiple scales. |
| **Unsupervised registration** | | | | | |
| Zhao et al., 2019b | 2019 | 3D liver CT image registration | *VTN:* several cascaded subnetworks | Private data, LITS, MICCAI'07 challenge | (1) Cascading the registration subnetworks to achieve better performance in registering largely displaced images; (2) Proposing invertibility loss for better accuracy. |
| Kim et al., 2019 | 2019 | 3D multiphase liver CT image registration | Based on *VoxelMorph* (Balakrishnan et al., 2018) | Private dataset | Performing unsupervised registration with cycle-consistency (Zhu et al., 2017). |
| Balakrishnan et al., 2018 | 2018 | 3D brain MRI registration | *VoxelMorph:* UNet-based network and STN | 8 public datasets (e.g. ADNI) | Formulating 3D image registration as a parametric function solving it without requiring supervised information. |

| Balakrishnan et al., 2019 | 2019 | 3D brain MRI registration | An extension of *VoxelMorph* | 8 public datasets (e.g. ADNI) | Extending *VoxelMorph* by leveraging auxiliary segmentation information (anatomical segmentation maps). |
|---|---|---|---|---|---|
| de Vos et al., 2017 | 2017 | 2D cardiac cine MR image registration | *DIRNet*: ConvNet and STN | Sunnybrook Cardiac Data | The first deep learning-based framework for end-to-end unsupervised deformable image registration. |
| de Vos et al., 2019 | 2019 | 3D cardiac cine MRI and chest CT registration | *DLIR*: stack of multiple CNNs | Sunnybrook Cardiac Data, NLST, etc. | (1) Extending *DIRNet* to 3D scenarios; (2) Introducing a multi-stage registration architecture by stacking multiple CNNs. |
| Fan et al., 2019b | 2019 | 3D brain MRI and multi-modal CT-MR pelvic image registration | GAN-based registration framework | LPBA40, IBSR18, CUMC12, MGH10, and private data | (1) Using the discriminator of GAN to implicitly learn an adversarial similarity to determine the voxel-to-voxel correspondence; (2) The proposed framework applies to both mono-modal and multi-modal registration. |

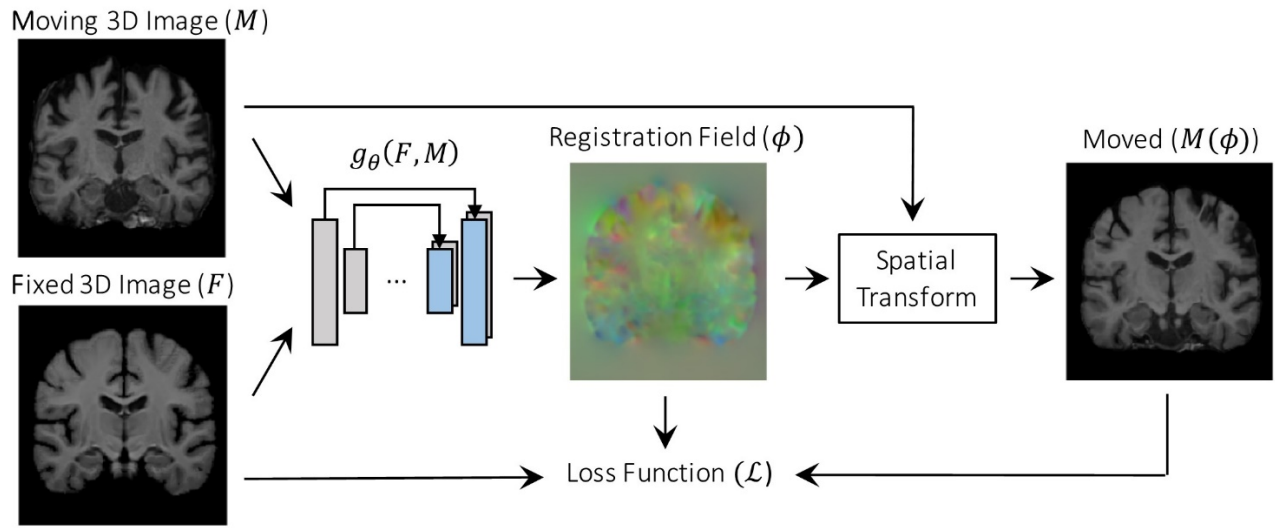

Figure 7. VoxelMorph (Balakrishnan et al., 2018).

# 4. DISCUSSIONS

## 4.1. Toward better combinations of deep learning and medical image analysis

### 4.1.1. On the task-specific perspective

The progress of medical image analysis using deep learning follows a lagging but similar timeline to computer vision. However, due to the difference between medical images and natural images, a direct use of methods from computer vision may not yield satisfactory results. In order to achieve good performance, challenges unique to medical imaging tasks need to be addressed. For the **classification** task, the key to success lies in extracting highly discriminative features with respect to certain classes. This is relatively easy for domains with large inter-class variance (e.g., accuracies on many public chest X-ray datasets often exceed 90%), but it can be difficult for domains with high inter-class similarity. For example, the performance of mammogram classification is not so good overall (e.g., 70~80% accuracies are commonly seen on private datasets), since discriminative features for breast tumors are difficult to capture in the presence of overlapping, heterogeneous fibroglandular tissues (Geras et al., 2019). The notion of *fine-grained visual classification (FGVC)* (Yang et al., 2018), which aims at identifying subtle differences between visually similar objects, might be suited for learning distinctive features given high inter-class similarity. But note that, benchmark FVGC datasets were purposely collected to make all the image samples unanimously exhibit high inter-class similarity. As a result, approaches developed and evaluated on these datasets may not be readily applicable to medical datasets, where only a certain fraction rather than all the images exhibit high inter-class similarity. Nonetheless, we believe FVGC methods, if modified appropriately, will be valuable to learning feature representations with high discriminative power in medical image classification. Other possible ways to enhance features' discrimination power include the use of attention modules, local and global features, domain knowledge, etc.

Medical object **detection** is more complicated than classification as can be seen from the process of bounding box prediction. Naturally, detection faces the challenges inherent to classification. Meanwhile, there exist additional challenges, especially the detection of small-scale objects (e.g., small lung nodules) and class imbalance. One-stage detectors typically perform comparably well as two-stage detectors in detecting large objects but struggles more in detecting small objects. Existing studies show that using multi-scale features can greatly alleviate this issue both in one-stage and two-stage detectors. A simple yet effective approach is *featurized image pyramids* (Liu et al., 2020b), where features are extracted from multiple scales of the same image independently. This method can help enlarge small objects to achieve better performance but is computationally expensive and slow. Nonetheless, it is suitable to medical detection tasks with no requirement of fast speed. Another useful but much faster approach is *feature pyramids*, which utilizes multi-scale feature maps from different convolutional layers. Although there exist various ways to build feature pyramids, a rule of thumb is that it is necessary to fuse strong, high-level semantics with high-resolution feature maps. This plays an important role in detecting small objects, as shown by FPN (Lin et al., 2017a).

Class imbalance arises from the fact that detectors need to evaluate a huge number of candidate regions, but only a few contain objects of interest. In other words, class balance is severely skewed toward negative examples (e.g., background regions), most of which are easy negatives. The presence of large amounts of easy negatives can overwhelm the training process, leading to bad detection results. Two-stage detectors can handle this class imbalance issue much better than one-stage detectors, because most negative proposals are filtered out at the region proposal stage. In terms of one-stage detectors, recent studies show that abandoning the dominant use of anchor boxes can largely alleviate class imbalance (Duan et al., 2019). However, most approaches adopted in medical object detection are still anchor-based. In the near future, we expect to see more explorations of anchor-free, one-stage detectors in medical object detection.

Medical image **segmentation** combines challenges in classification and detection. Just like detection, class imbalance is a common issue across 2D and 3D medical segmentation tasks. Another similar challenge is the segmentation of small-sized lesions (e.g., MRI multiple sclerosis) and organs (e.g., pancreas from abdominal CT scans). Also, these two challenges often appear intertwined. These issues have been largely alleviated by adapting metrics/losses to evaluate the segmentation performance, such as Dice coefficient

(Milletari et al., 2016), generalized Dice (Sudre et al., 2017), the integration of focal loss (Abraham and Khan, 2019), etc. However, these metrics are region-based (i.e., segmentation errors are computed in a pixel-wise manner). This can lead to a loss of valuable information regarding structures, shapes, and contours that are important to diagnosis/prognosis in later stages. Therefore, we believe it is necessary to develop non-region-based metrics that can provide complementary information to region-based metrics for better segmentation performance. Currently only a few studies exist in this direction (Kervadec et al., 2019). We expect to see more in the future.

In addition, strategies such as incorporating local and global context, attention mechanisms, multi-scale features, and anatomical cues are generally beneficial to increasing segmentation accuracy for both large and small objects. Here we want to emphasize the great potentials of Transformers due to their strong capability of modeling long-range dependencies. Although long-range dependencies are helpful to achieving accurate segmentation, a majority of CNN-based methods do not explicitly focus on this aspect. There are roughly two types of dependencies, namely intra-slice dependency (pixel relationships within a CT or MRI slice) and inter-slice dependency (pixel relationships between CT or MRI slices) (Li et al., 2020e). Recent studies show that Transformer-based approaches are powerful in both cases (Chen et al., 2021b; Valanarasu et al., 2021). Applications of Transformers for medical image segmentation especially 3D are still in the initial stage, and more works in this trial are likely to emerge soon.

Medical image **registration** significantly differs from previous tasks because its purpose is to find the pixel-wise or voxel-wise correspondence between two images. One unique challenge is associated with the difficulty in acquiring reliable ground truth registrations, which are either synthetically generated or produced by conventional registration algorithms. Unsupervised methods have shown great promise in solving this issue. However, many unsupervised registration frameworks (e.g. de Vos et al., 2019) are composed of multiple stages to register images in a coarse-to-fine manner. Despite the good performance, multi-stage frameworks can increase computational complexity and make training difficult. It would be desirable to develop registration frameworks that have as few stages as possible and can be trained end to end.

### 4.1.2. On the perspective of different learning paradigms

Although deep learning has brought about huge successes across different tasks in the context of radiological image analysis, the further performance improvement is majorly hurdled by the requirement for large amounts of annotated datasets. Supervised transfer learning can greatly alleviate this issue, by initializing the model's weights (for the target task) with the weights of the model that is pre-trained on relevant/irrelevant datasets (e.g. ImageNet). Besides the widely used transferring learning, there are two possible directions: (1) utilizing GAN model to enlarge the labeled dataset; (2) utilizing the self-supervised and semi-supervised learning models to exploit the information underlying vast unlabeled medical images.

**GAN** has shown great promise in medical image synthesis and semi-supervised learning; but one challenge is how to build a strong connection between GAN's generator and the target task (e.g., classifier, detector, segmentor). The lack of such connection may cause a subtle performance boost as compared to the conventional data augmentation (e.g., rotation, rescale, and flip) (Wu et al., 2018a). The connection between the generator and classifier can be strengthened by utilizing semi-supervised GAN, in which the discriminator was modified to serve as a classifier (Salimans et al., 2016). Several training strategies can also be employed: identifying a "bad" generator that can significantly contribute to good semi-supervised classification (Dai et al., 2017); jointly optimizing the triple components of a generator, a discriminator, and a classifier (Li et al., 2017). It is meaningful to explore new ways that can effectively set up connections between the generator and a specific medical image task for a better performance. Additionally, GAN usually needs at least thousands of training examples to converge, which limits its applicability on small medical datasets. This challenge can be partially addressed by using classic data augmentation for adversarial learning (Frid-Adar et al., 2018a; Frid-Adar et al., 2018b). Further, if there exist relatively large amounts of medical images that share structural, textural, and semantic similarities with the target dataset, pre-training generators and/or discriminators may facilitate faster convergence and better performance (Rubin et al., 2019). Meanwhile, some recent novel augmentation mechanisms, such as the differentiable augmentation (Zhao et al., 2020) and adaptive discriminator augmentation (Karras et al., 2020) have enabled GAN to effectively generate high-fidelity images under data-limited conditions, but they have not been applied to any medical image analysis tasks. We anticipate that these new methods can also demonstrate promising performance in future studies of the medical image analysis field.

**Self-supervision** can be constructed by either pretext tasks or contrastive learning, but the latter seems to be a more promising research direction. This is because, on one hand directly using pretext tasks (e.g. jigsaw puzzle) from computer vision is typically not adequate to ensure learning robust feature representations for radiological images. On the other hand, designing novel pretext tasks can be difficult, which demands delicate manipulation. Instead of designing various pretext tasks, self-supervised contrastive learning trains the network to capture meaningful feature representations by forcing them to be invariant to different augmented views, which can potentially outperform supervised transfer learning on different downstream tasks, such as medical image classification and segmentation. Despite the encouraging performance of self-supervised contrastive learning, its applications in radiological image analysis are still at the exploratory stage, and how to make appropriate use of this new learning paradigm is a difficult problem. To unleash its potential, here we provide our suggestions from the following three aspects. (1) Harness the benefits of contrastive learning and supervised learning. Observing from the exiting studies, we find a majority adopt two separate steps for medical image analysis: contrastive pre-training on unlabeled data and supervised fine-tuning with labeled data. At the pre-training stage, most studies are reliant on relatively large, unlabeled datasets to ensure learning high-quality, transferrable features, which can yield superior performance after being tuned using limited labeled data. However, the reliance on large unlabeled data could be problematic in tasks lacking large amounts of unlabeled data. To expand the application scope, learning high-quality feature presentations with less unlabeled data would be desirable. One possible approach is unifying the previously mentioned two separate steps into one so that the label information can be leveraged in contrastive learning. This is somewhat reminiscent of semi-supervised learning that simultaneously utilizes unlabeled and labeled data to achieve better performance. More concretely, class labels can be used to guide constructing positive and negative pairs in a more compact manner by pushing images from the same class to be more closely aligned in the lower-dimensional representation space (Khosla et al., 2020). Features learned in this way should require less unlabeled data and be less redundant than features learned solely through self-supervised learning (i.e., without any class labels). (2) Take into account certain properties of contrastive learning for better performance. For example, one study proves that contrastive learning benefits more from large blocks of similar points than pairs (Saunshi et al., 2019). This heuristic may be well suited to learning transferrable features from 3D CT and MRI volumes exhibiting consecutive anatomical similarity. (3) Customize data augmentation strategies for downstream tasks that are sensitive to augmentation. The composition of different data augmentation strategies proves critical to learning representative features in most existing contrastive learning frameworks. For instance, SimCLR applies three types of transformations to unlabeled images, namely random cropping, color distortions, and Gaussian blur (Chen et al., 2020a). However, some commonly used augmentation techniques may not be applicable to medical images. In radiology, where most images are presented in grayscales, the color distortion strategy is likely not suitable. Also, in cases where fine-grained details of unlabeled medical image carry important information, applying Gaussian blur may ruin the detailed information and degenerate the quality of feature representations during the pre-training stage. Therefore, it is important to choose appropriate data augmentation strategies to ensure satisfactory downstream performance. In addition, self-supervised contrastive pre-training is currently impeded by the high computing complexity of large models (e.g., ResNet-50 (4×), ResNet-152 (2×)), which require a large group of multi-core TPUs (Chen et al., 2020a). Therefore, it should be an important direction to develop novel models or training strategies to enhance the computing efficiency. For example, Reed et al. (2022) proposed a hierarchical pre-training strategy to make the self-supervised pre-training process converge up to 80× faster with an improved accuracy across different tasks.

Like self-supervised contrastive learning, recent **semi-supervised** methods such as FixMatch (Sohn et al., 2020) heavily rely on advanced data augmentation strategies to achieve good performance. To facilitate the applications of semi-supervised learning in medical image analysis, it is necessary to develop appropriate augmentation policies in a *dataset-driven* and/or *task-driven* manner. Being "dataset-driven" means finding the optimal augmentation policy for a specific dataset of interest. In the past, this was not easy to achieve due to the extremely very large size of the parameter search space (e.g., $10^{34}$ possible augmentation policies as shown by Cubuk et al. (2020)). Recently, automated data augmentation strategies like *RandAugment* (Cubuk et al., 2020) have been proposed to significantly reduce the search space. However, the concept of automated augmentation remains largely unexplored in medical image analysis. Being "task-driven" means finding suitable augmentation strategies for a specific task (e.g., MRI prostate segmentation) that have several datasets. This could be regarded as the extension of dataset-driven augmentation and thus is more challenging, but it can help algorithms developed on one dataset generalize better to other dataset(s) of the same task.

Another issue is the potential performance degradation caused by violation of the underlying assumption of semi-supervised learning – labeled and unlabeled data are from the same distribution. Indeed, distribution mismatch is a common problem when semi-supervised methods are applied for medical image analysis. Consider the following example: in the task of segmenting COVID-19 lung infections from CT slices, say you have a set of labeled CT volumes containing a relatively balanced number of infected and non-infected slices, while the unlabeled CT volumes available may contain no or just a few infected slices. Or the unlabeled CT images contain not only COVID-19 infections but also some other disease class(es) (e.g., tuberculosis) that are absent from the labeled images. What will happen if the distribution of unlabeled data mismatches with the distribution of labeled data? Exiting studies suggest this will cause the performance of semi-supervised methods to degrade drastically, sometimes even worse than that of a simple supervised baseline (Oliver et al., 2018; Guo et al., 2020). Therefore, it is necessary to adapt semi-supervised algorithms to be tolerant of the distribution mismatch between labeled and unlabeled medical data. As a related field, "domain adaption" may provide insights for achieving this goal.

### 4.1.3. Finding better architectures and pipelines

The continuing success of deep learning in medical image analysis originates from not only different learning paradigms (unsupervised, semi-supervised) but also, maybe to a larger extent, the architectures/models proposed over time. Looking back, we find non-trivial improvements are closely related to the progress of "architectures", and examples include AlexNet (Krizhevsky et al., 2012), residual connections (He et al., 2016), skip connections (Ronneberger et al., 2015), self attention (Vaswani et al., 2017) etc. "Given this progression history, it is certainly possible that a better neural architecture can by itself overcome many of the current limitations", as pointed out by Yuille and Liu (2021). We disucuss two aspects that may be helpful to finding better architectures. First, biologically and cognitively inspired mechanisms will continue playing an important role in architecture designing. Deep learning neural networks were originally inspired by the architecture of cerebal context. In recent years the concept of attention, which was inspired by primates' visual attention mechanisms, has been successfully used in NLP and computer vision to make models focus on important parts of input data, leading to superior performance. A preeminent example are the family of Transformers based on self attention (Vaswani et al., 2017). Transformer-based architectures are better at capturing global/long-range dependencies between input and output sequences than mainstream models based on CNNs. Also, inductive biases inherent to CNNs (e.g., translation equivarance and locality) are much less in Transformers (Dosovitskiy et al., 2020). Aside from the attention mechanisms, many other biological or cognitive mechanisms, such as dynamic hierarchies in human language, one-shot learning of new objects and concepts without gradient descent, etc (Marblestone et al., 2016), may provide inspirations for designing more powerful achitectures. Second, automatic architecture engineering may shed light on developing better architectures. Currently employed architecutres mostly come from human experts, and the designing process is iterative and prone to errors. Partially for this reason, models used for medical image anlaysis are primarily adapted from models developed in computer vision. To avoid the need of manual designing, reserachers have proposed to automate architecutre engineering, and one related field is *neural architecture search* (NAS) (Zoph and Le, 2016). However, most exisiting studies of NAS are confined within image classification (Elsken et al., 2019), and truly revolutionary models that can bring fundamental changes have not come out of this process (Yuille and Liu, 2021). Nonetheless, NAS is still a direction worthy exploration.

At a broader level, pipelines with automated configuration capabilities would be desirable. Although architecture engineering still faces many difficulties, developing automatic pipelines, which are capable of automatically configuring its subcomponents (e.g., choosing and adapting an appropriate architecture among the exisiting ones) to achieve better performance, will be beneficial to radiolgical image analysis. At present, deep leanring based pipelines typically involve several interdepedent subcomponents such as image preprocessing and post-processing, adapting and training a network architecture, selecting appropriate losses, data augmentation methods, etc. But the design choices are often too many for experimenters to manually figure out an optimal pipeline. Moreover, a high-performing pipeline configured for a dataset (e.g., CT images from one hospital) of a specific task may perform badly on anther dataset (e.g., CT images from a different hospital) of the same task. Therefore, pipelines that can automatically configure their subcomponents are needed to speed up empirical design. Examples falling in this scope include NiftyNet (Gibson et al., 2018b), a modular pipeline for different medical applications, and nnU-Net (Isensee et al., 2021) specifically for medical image segmentation. We expect more research will be coming out of this track.

### 4.1.4. Incorporating domain knowledge

Domain knowledge, which is an important aspect but sometimes overlooked, can provide insights for developing high-performing deep learning algorithms in medical image analysis. As mentioned previously, most models used in medical vision are adapted from models developed for natural images; however, medical images are generally more difficult to handle due to unique challenges (e.g., high inter-class similarity, limited size of labeled data, label noise). Domain knowledge, if used appropriately, helps alleviate these issues with less time and computation costs. It is relatively easy for researchers with strong deep learning background to utilize *weak* domain knowledge, such as anatomical information in MRI and CT images (Zhou et al., 2021; Zhou et al., 2019a), multi-instance data from the same patient (Azizi et al., 2021), patient metadata (Vu et al., 2021), radiomic features, and text reports accompanying images (Zhang et al., 2020a). On the other hand, we observe it can be more difficult to effectively incorporate *strong* domain knowledge that radiologists are familiar with. One example is breast cancer identification from mammograms. For each patient, four mammograms are available, including two cranio-caudal (CC) and two medio-lateral oblique (MLO) view images of left (L) and right (R) breasts. In clinical practice, the bilateral difference (e.g. LCC vs. RCC) and unilateral correspondence (e.g. LCC and LMLO) serve as important cues for radiologists to detect suspicious regions and determine malignancy. Currently there exist few methods that can reliably and accurately to utilize this expert knowledge. Therefore, more research efforts are needed to maximize the use of strong domain knowledge.

### 4.2. Toward large-scale applications of deep learning in clinical settings

Deep learning, despite being intensively used for analyzing medical images in academia and industrial research institutions, has not made that significant impact as we expected in clinical practice. This is clearly reflected in the early stages of fighting against COVID-19, the first global pandemic falling in the era of deep learning. Due to its widespread medical, social, and economic consequences, this pandemic, to a large extent, can be regarded as a big test for examining the current status of deep learning algorithms in clinical translation. Soon after the outbreak, researchers around the world applied deep learning techniques to analyze mainly chest X-rays and CT images from patients with suspected infection, aiming at accurate and efficient diagnosis/prognosis of the disease. To this end, numerous deep learning and machine learning based approaches were developed. However, after systematically reviewing over 200 prediction models from 169 studies that were published up to 1 July 2020, Wynants et al. (2020) concluded that all these models were of high or unclear risk of bias, and thus none of them were suitable for clinical use – either moderate or excellent performance was reported by each model; however, the optimistic results were highly biased due to model overfitting, inappropriate evaluation, use of improper data sources, etc. Similar conclusion was drawn in another review paper (Roberts et al., 2021) – after reviewing 62 studies that were selected from 415 studies the authors concluded that, because of methodological flaws and/or underlying biases, none of the deep learning and machine learning models identified were clinically applicable to the diagnosis/prognosis of COVID-19.

Going beyond the example of COVID-19, the high-risk bias of deep learning approaches is indeed a recurring concern across different medical image analysis tasks and applications (Nagendran et al., 2020), which has severely limited deep learning's potential in clinical radiology. Although quantifying the underlying bias is difficult, it can be reduced if handled appropriately. In the following we summarize three major aspects that could lead to biased results and provide our recommendations.

### 4.2.1. Image datasets

Data forms the basis of deep learning. In medical vision, medical image datasets with increasingly larger size (e.g. usually at least several hundred images) have been or are being developed to facilitate training and testing new algorithms. One notable example is the yearly MICCAI challenges where benchmark datasets for different diseases (e.g. cancer) are released, greatly promoting the progress of medical vision. However, we need to be cautious about the potential biases caused by using a single public dataset alone – as the whole community strive for achieving state of the art performance, community-wide overfitting is likely to exist on this dataset (Roberts et al., 2021). This problem has been recognized by many researchers, so it is common to see several public datasets and/or private dataset(s) are used to test a new algorithm's performance more comprehensively. In this way the community-wide bias is reduced but not to the extent of large-scale clinical applications.

The community-wide bias can be further lowered by incorporating additional data to train and test models. One direct way to introduce new data, of course is data curation, i.e., continually creating large, diverse datasets via collective work with experts. Different from this track, we recommend a less direct but effective way – integrating scattered private datasets as ethical and law regulations permit. The medical image analysis community might have the overall impression that large, representative, labeled data seems always lacking. This is only partially true, though. Due to time and cost constraints, it is true that many established public datasets have limited size and variety. On the other hand, rich medical image sources (labeled and unlabeled) of different sizes and difficulty levels already exist but inconveniently "in the form of isolated islands" (Yang et al., 2019). Because of factors such as privacy protection and political intricacy, most existing data sources are kept private and scattered in different institutions across different countries. Thus, it would be desirable to exploit the unified potentials of private datasets and even personal data without comprising patients' privacy. A promising approach to achieving this goal is *federated learning* (Li et al., 2020f), which allows models to securely access sensitive data. Federated learning can train deep learning algorithms collaboratively on multi-institutional data without exchanging data among participating institutions (Rieke et al., 2020). Although this technology is accompanied by new challenges, it facilitates learning less biased, more generalizable, more robust, and better-performed algorithms that would better meet the needs of clinical applications.

### 4.2.2. Performance evaluation

Most research papers in medical image analysis report models' performance via commonly used metrics, for example, accuracy and AUC for classification tasks, and Dice coefficient for segmentation tasks. While these metrics can easily quantify the technical performance of presented approaches, they often fail to reflect clinical applicability. Ultimately, clinicians care about whether the use of algorithms would bring about a beneficial change in patient care, rather than the performance gains reported in papers (Kelly et al., 2019). Therefore, aside from applying necessary metrics, we believe it is important for research teams to collaborate with clinicians for algorithms appraisal.

We simply mention two possible directions as to establishing collaborative evaluation. First, involve clinicians into viewpoints sharing of open clinical questions, paper writing, and even the peer review process of conferences and journals. For example, the *Machine Learning for Healthcare* (MLHC) conference provides a research track and clinical track for members from separated communities to exchange insights. Second, measure if the performance and/or efficiency of clinicians can be improved with the assistance of deep learning algorithms. Utilizing model results as a "second opinion" to facilitate clinicians' final interpretation has been explored in some studies. For instance, in the task of predicting breast cancer from mammograms, McKinney et al. (2020) evaluated the complementary role of deep learning model. They found that the model could correctly identify many cancer cases missed by radiologists. Furthermore, in the "double-reading process" (standard practice in UK), the model significantly reduced the second reader's workload while maintaining a comparable performance to the consensus opinion.

### 4.2.3. Reproducibility

The quick progress of computer vision is closely related to the research culture that advocates reproducibility. In medical image analysis, more and more researchers choose to make their code publicly available, and this greatly helps avoid duplication of effort. More importantly, good reproducibility can help deep learning algorithms gain more trust and confidence from a wide population (e.g., researchers, clinicians), which is beneficial to large-scale clinical applications. To make the results more reproducible, we suggest paying extra attention to describing data selection in papers. It is not uncommon to see that different studies select different subsets of samples from the same public dataset. This could increase the difficulty of reproducing results stated in the paper. In a case study on lung nodule classification, Baltatzis et al. (2021) demonstrated that specific choices of data turn out to be favorable to proving the proposed models' superiority. Advanced models with bells and whistles may underperform simple baselines if data samples are changed. Therefore, it is necessary to clearly state the data selection process to make the results more reproducible and convincing.

In conclusion, deep learning is a fast-developing technology, and has produced promising potential in broad medical image analysis fields including disease classification, segmentation, detection, and image registration. Despite of significant research progress, we are still facing many technical challenges or pitfalls (Roberts et al., 2021) to develop deep learning based CAD schemes that can achieve high scientific rigor. Therefore, more research efforts are needed to overcome these pitfalls before the deep learning based CAD schemes can be commonly accepted by clinicians.

## 5. DECLARATION OF COMPETING INTEREST

The authors declare that they have no known competing financial interests or personal relationships that could have appeared to influence the work reported in this paper.

## 6. ACKNOWLEDGMENTS

The authors gratefully acknowledge the following research support: Grant P20GM135009 from National Institute of General Medical Sciences, National Institutes of Health; Stephenson Cancer Center Team Grant funded by the National Cancer Institute Cancer Center Support Grant P30CA225520 awarded to the University of Oklahoma Stephenson Cancer Center.